\documentclass{article}
\usepackage[dvipsnames]{xcolor}

\usepackage[final,nonatbib]{neurips_2020}

\usepackage[square,sort,comma,numbers]{natbib}
\usepackage[utf8]{inputenc} %
\usepackage[T1]{fontenc}    %
\usepackage{hyperref}       %
\usepackage{url}            %
\usepackage{booktabs}       %
\usepackage{amsfonts}       %
\usepackage{nicefrac}       %
\usepackage{microtype}      %
\usepackage{graphicx}
\usepackage{subcaption}
\usepackage{pgf}
\usepackage{tikz}
\usepackage{wrapfig}
\usepackage{multirow}
\usepackage{tabularx}
\usetikzlibrary{arrows,arrows.meta}
\usepackage{amsmath,amsfonts,amssymb,bm,xspace,nicefrac}
\usepackage{array}
\usepackage{pifont} %

\newcommand{\PreserveBackslash}[1]{\let\temp=\\#1\let\\=\temp}
\newcolumntype{C}[1]{>{\PreserveBackslash\centering}p{#1}}
\newcolumntype{R}[1]{>{\PreserveBackslash\raggedleft}p{#1}}
\newcolumntype{L}[1]{>{\PreserveBackslash\raggedright}p{#1}}

\newcommand{\CumulativeIntensity}{\Lambda^*}

\newcommand{\History}{\mathcal{H}}

\newcommand{\obs}{\vo}
\newcommand{\intensity}{\lambda}
\newcommand{\sintensity}{\lambda^*}
\newcommand{\mapF}{\mF}
\newcommand{\mapLa}{\bm{\Lambda}}
\newcommand{\mapPh}{\bm{\Phi}}
\newcommand{\vpsi}{\bm{\psi}}

\newcommand{\name}{TriTPP\xspace}
\newcommand{\elbo}{\operatorname{ELBO}}
\newcommand{\temp}{\gamma}
\newcommand\clip[1]{#1}
\newcommand\ext[1]{\tilde{#1}}
\newcommand\hl[1]{\textcolor{BrickRed}{#1}}

\newcommand{\Categorical}{\operatorname{Categorical}}

\newcommand{\Exponential}{\operatorname{Exponential}}

\newcommand\second[1]{\underline{#1}}
\newcommand{\dz}{\tilde{p}}

\def\Figref#1{Figure~\ref{#1}}

\def\Appref#1{Appendix~\ref{#1}}
\def\Secref#1{Section~\ref{#1}}

\def\eqref#1{equation~\ref{#1}}
\def\Eqref#1{Equation~\ref{#1}}

\def\1{\bm{1}}

\def\vone{{\bm{1}}}
\def\vlambda{{\bm{\lambda}}}

\def\vtheta{{\bm{\theta}}}
\def\vpi{{\bm{\pi}}}
\def\vsigma{{\bm{\sigma}}}

\def\vj{{\bm{j}}}

\def\vm{{\bm{m}}}

\def\vo{{\bm{o}}}

\def\vs{{\bm{s}}}
\def\vt{{\bm{t}}}
\def\vu{{\bm{u}}}

\def\vx{{\bm{x}}}

\def\vz{{\bm{z}}}

\def\mA{{\bm{A}}}
\def\mB{{\bm{B}}}
\def\mC{{\bm{C}}}
\def\mD{{\bm{D}}}

\def\mF{{\bm{F}}}
\def\mG{{\bm{G}}}

\def\mI{{\bm{I}}}

\DeclareMathAlphabet{\mathsfit}{\encodingdefault}{\sfdefault}{m}{sl}
\SetMathAlphabet{\mathsfit}{bold}{\encodingdefault}{\sfdefault}{bx}{n}

\def\gD{{\mathcal{D}}}

\def\gO{{\mathcal{O}}}

\newcommand{\E}{\mathbb{E}}

\newcommand{\R}{\mathbb{R}}

\newcommand{\sigmoid}{\sigma}

\DeclareMathOperator*{\argmax}{arg\,max}

\title{Fast and Flexible Temporal Point Processes\\ with Triangular Maps}

\author{%
  Oleksandr Shchur, Nicholas Gao, Marin Bilo\v{s}, Stephan G\"{u}nnemann\\
  Technical University of Munich, Germany\\
  \texttt{\{shchur,gaoni,bilos,guennemann\}@in.tum.de} \\
}

\begin{document}

\maketitle

\begin{abstract}
Temporal point process (TPP) models combined with recurrent neural networks provide a powerful framework for modeling continuous-time event data.
While such models are flexible, they are inherently sequential and therefore cannot benefit from the parallelism of modern hardware.
By exploiting the recent developments in the field of normalizing flows, we design \name --- a new class of non-recurrent TPP models, where both sampling and likelihood computation can be done in parallel.
\name matches the flexibility of RNN-based methods but permits orders of magnitude faster sampling.
This enables us to use the new model for variational inference in continuous-time discrete-state systems.
We demonstrate the advantages of the proposed framework on synthetic and real-world datasets.
\let\thefootnote\relax\footnotetext{Code and datasets are available under \url{www.daml.in.tum.de/triangular-tpp}}
\end{abstract}

\section{Introduction}

Temporal data lies at the heart of many high-impact machine learning applications.
Electronic health records, financial transaction ledgers and server logs contain valuable information.
A common challenge encountered in all these settings is that both the number of events and their times are variable.
The framework of temporal point processes (TPP) allows us to naturally handle data that consists of variable-number events in continuous time.
Du et al.\ \citep{du2016rmtpp} have shown that the flexibility of TPPs can be improved by combining them with recurrent neural networks (RNN).
While such models are expressive and can achieve good results in various prediction tasks, they are poorly suited for sampling:
sequential dependencies preclude parallelization.
We show that it's possible to overcome the above limitation and design flexible TPP models without relying on RNNs.
For this, we use the framework of triangular maps \citep{jaini2019sum} and recent developments in the field of normalizing flows \citep{durkan2019neural}.

Our main contributions are: \textbf{(1)}
We propose a new parametrization for several classic TPPs. This enables efficient parallel likelihood computation and sampling, which was impossible with existing parametrizations.
\textbf{(2)} We propose \name --- a new class of non-recurrent TPPs. 
\name matches the flexibility of RNN-based methods, while allowing orders of magnitude faster sampling.
\textbf{(3)} We derive a differentiable relaxation for non-differentiable sampling-based TPP losses.
This allows us to design a new variational inference scheme for Markov jump processes.

\section{Background}
\textbf{Temporal point processes (TPP)} \citep{daley2003introduction}
are stochastic processes that model the distribution of discrete events on some continuous time interval $[0, T]$.
A realization of a TPP is a \emph{variable-length} 
sequence of strictly increasing arrival times $\vt = (t_1, \dots, t_N), t_i \in [0, T]$.
We make the standard assumption and focus our discussion on regular finite TPPs \citep{daley2003introduction}.
One way to specify such a TPP is by using the (strictly positive) conditional intensity function $\sintensity(t) := \intensity(t | \History_t)$ that defines
the rate of arrival of new events given the history $\History_t = \{t_j : t_j < t\}$.
The $*$ symbol reminds us of the dependence on the history \citep{rasmussen2011temporal}.
Equivalently, we can consider the \emph{cumulative} conditional intensity $\CumulativeIntensity(t):= \Lambda(t | \History_{t}) = \int_{0}^{t} \sintensity(u) du$, also known as the compensator.\footnote{For convenience, we provide a list of abbreviations and notation used in the paper in \Appref{app:notation}.}
We can compute the likelihood of a realization $\vt$ on $[0, T]$ as
\begin{align}
    \label{eq:tpp-likelihood}
    p(\vt)
    &= \left(\prod_{i=1}^{N} \sintensity(t_i)\right) \exp \left( -\int_{0}^{T} \sintensity(u) du\right)
    = \left(\prod_{i=1}^N \frac{\partial}{\partial t_i} \CumulativeIntensity(t_i)\right) \exp \left( -\CumulativeIntensity(T) \right) 
\end{align}
For example, we can use a TPP to model the online activity of a user in a 24-hour interval.
In this case, each realization $\vt$ could correspond to the timestamps of the posts by the user on a specific day.

\textbf{Triangular maps} \citep{jaini2019sum} 
provide a framework that connects autoregressive models, normalizing flows and density estimation.
Bogachev et al.\ \citep{bogachev2005triangular} have shown that any density $p(\vx)$ on $\R^N$ can be equivalently represented 
by another density $\dz(\vz)$ on $\R^N$ and an increasing differentiable triangular map $\mapF = (f_1, \dots, f_N) : \R^N \to \R^N$ that pushes forward $p$ into 
$\dz$.\footnote{Note that some other works instead define $\mapF$ as the map that pushes the density $\dz(\vz)$ into $p(\vx)$.} %
A map $\mapF$ is called triangular if each component function $f_i$ depends only on $(x_1, \dots, x_i)$ and is an increasing function of $x_i$.
Intuitively, we can think of $\mapF$ as converting a random variable $\vx \sim p$ into a random variable $\vz := \mapF(\vx)$ with a density $\dz$. 
We can compute the density $p(\vx)$ using the change of variables formula
\begin{align}
    \label{eq:change-of-vars}
    p(\vx) &= \left|\operatorname{det} J_{\mapF}(\vx) \right| \dz\left(\mapF(\vx)\right) 
    = \left( \prod_{i=1}^N \frac{\partial}{\partial x_i} f_i(x_1, \dots, x_i) \right) \dz\left(\mapF(\vx)\right)
\end{align}
where $\operatorname{det} J_{\mapF}(\vx)$ is the Jacobian determinant of $\mapF$ at $\vx$.
Here, we used the fact that $J_{\mapF}(\vx)$ is a positive-definite lower-triangular matrix.
To specify a complex density $p(\vx)$,
we can pick some simple density $\dz(\vz)$ and learn the triangular map $\mapF$ that pushes $p$ into $\dz$.
It's important that $\mapF$ and its Jacobian determinant can be evaluated efficiently if we are learning $p(\vx)$ via maximum likelihood.
We can sample from $p(\vx)$ by applying the inverse map $\mapF^{-1}$ to the samples drawn from $\dz(\vz)$.
Note that $\mapF^{-1} : \R^N \to \R^N$ is also an increasing differentiable triangular map.
Fast computation of $\mapF^{-1}$ is important when learning $p(\vx)$ via sampling-based losses (e.g., in variational inference).

\section{Defining temporal point processes using triangular maps}
\label{sec:tri-tpp}
We can notice the similarity between the right-hand sides of Equations \ref{eq:tpp-likelihood} and \ref{eq:change-of-vars}, which seems to suggest some connection between TPPs and triangular maps.
Indeed, it turns out that triangular maps can also be used to specify densities of point processes.
Let $\vt = (t_1, \dots, t_N)$ be a realization of a TPP on $[0, T]$ with compensator $\CumulativeIntensity$ (i.e.\ with density $p(\vt)$).
The random time change theorem states that in this case $\vz = (\CumulativeIntensity(t_1), \dots, \CumulativeIntensity(t_N))$ 
is a realization of a homogeneous Poisson process (HPP) with unit rate on the interval $[0, \CumulativeIntensity(T)]$ 
\citep[Theorem 7.4.I]{daley2003introduction}\citep[Proposition 4.1]{rasmussen2011temporal} (\Figref{fig:density}).

The transformation $\mapF = (f_1, \dots, f_N) : \vt \mapsto \vz$ is an increasing triangular map. 
Each component function $f_i(\vt) = \Lambda(t_i | t_1, \dots, t_{i-1})$ only depends on $(t_1, \dots, t_i)$ and is increasing in $t_i$ since $\frac{\partial}{\partial t_i} \CumulativeIntensity(t_i) = \sintensity(t_i) > 0$.
The number $N$ of the component functions $f_i$ depends on the length of the specific realization $\vt$.
Notice that the term $\prod_{i=1}^N \frac{\partial}{\partial t_i} \CumulativeIntensity(t_i)$ in \Eqref{eq:tpp-likelihood} corresponds to the Jacobian determinant of $\mapF$.
Similarly, the second term, $\dz(\vz) = \dz(\mapF(\vt)) = \exp(-\CumulativeIntensity(T))$, corresponds to the density of a HPP with unit rate on $[0, \CumulativeIntensity(T)]$ for any realization $\vz$.
This demonstrates that all TPP densities (\Eqref{eq:tpp-likelihood}) correspond to increasing triangular maps (\Eqref{eq:change-of-vars}).
As for the converse of this statement, every increasing triangular map that is bijective on the space of increasing sequences defines a valid TPP (see \Appref{app:valid}).

Our main idea is to define TPP densities $p(\vt)$ by directly specifying the respective maps $\mapF$.
In \Secref{sec:tri-tpp-requirements}, we show how maps that satisfy certain properties allow us to efficiently compute density and generate samples.
We demonstrate this by designing a new parametrization for several established models in \Secref{sec:tri-tpp-nonar}.
Finally, we propose a new class of fast and flexible TPPs in \Secref{sec:tri-tpp-triangular}.

\subsection{Requirements for efficient TPP models}
\label{sec:tri-tpp-requirements}
\textbf{Density evaluation.}
The time complexity of computing the density $p(\vt)$ for various TPP models can be understood by analyzing the respective map $\mapF$.
For a general triangular map $\mapF : \R^N \to \R^N$, computing $\mapF(\vt)$ takes $\gO(N^2)$ operations.
For example, this holds for Hawkes processes with
\begin{wrapfigure}{r}{0.30\textwidth}
    \vspace{-7mm}
    \resizebox{0.3\textwidth}{!}{\input{figures/fig11.pgf}}
    \vspace{-10mm}
    \caption{Triangular map $\mapF(\vt) = (\CumulativeIntensity(t_1), ..., \CumulativeIntensity(t_N))$ is used for computing $p(\vt)$.}
    \label{fig:density}
    \resizebox{0.3\textwidth}{!}{\input{figures/fig12.pgf}}
    \vspace{-10mm}
    \caption{Sampling is done by applying $\mapF^{-1}$ to a sample $\vz$ from a HPP with unit rate.}
    \label{fig:sampling}
    \vspace{-5mm}
\end{wrapfigure}
arbitrary kernels \citep{hawkes1971spectra}.
If the compensator $\CumulativeIntensity$ has Markov property, the complexity of evaluating $\mapF$ can be reduced to $\gO(N)$ \emph{sequential} operations.
This class of models includes Hawkes processes with exponential kernels \citep{oakes1975markovian,dassios2013exact} and RNN-based autoregressive TPPs \citep{du2016rmtpp,omi2019fully,shchur2020intensity}.
Unfortunately, such models do not benefit from the parallelism of modern hardware.
Defining an efficient TPP model will require specifying a forward map $\mapF$ that can be computed in $\gO(N)$ \emph{parallel} operations.

\textbf{Sampling.}
As a converse of the random time change theorem, we can sample from a TPP density $p(\vt)$ by first drawing $\vz$ from an HPP on $[0, \CumulativeIntensity(T)]$ and applying the inverse map, $\vt = \mapF^{-1}(\vz)$~\citep{daley2003introduction}.
There are, however, several caveats to this method.
Not all parametrizations of $\mapF$ allow computing $\mapF^{-1}(\vz)$ in closed form.
Even if $\mapF^{-1}$ is available, its evaluation for most models is again sequential \citep{du2016rmtpp,dassios2013exact}.
Lastly, the number of points $N$ that will be generated (and thus $\CumulativeIntensity(T)$ for HPP) is not known in advance.
Therefore, existing methods typically resort to generating the samples one by one \citep[Algorithm 4.1]{rasmussen2011temporal}.
We show that it's possible to do better than this.
If the inverse map $\mapF^{-1}$ can be applied in parallel, we can produce large batches of samples $t_i$, and then discard the points $t_i > T$ (\Figref{fig:sampling}).
Even though this method may produce samples that are later discarded, it is much more efficient than sequential generation on GPUs (\Secref{sec:exp-scale}).

To summarize, defining a TPP efficient for both density computation and sampling requires specifying a triangular map $\mapF$, such that both $\mapF$ and its inverse $\mapF^{-1}$ can be evaluated analytically in $\gO(N)$ \emph{parallel} operations.
We will now show that maps corresponding to several classic TPP models can be defined to satisfy these criteria.

\subsection{Fast temporal point process models}
\label{sec:tri-tpp-nonar}
\textbf{Inhomogeneous Poisson process (IPP)} \citep{daley2003introduction}
is a TPP whose conditional intensity doesn't depend on the history, $\Lambda(t | \History_t) = \Lambda(t)$.
The corresponding map is $\mapF = \mapLa$, where $\mapLa$ simply applies the function $\Lambda : [0, T] \to \R_+$ elementwise to the sequence $(t_1, ..., t_N)$.

\textbf{Renewal process (RP)} \citep{cox1962renewal} is a TPP where each inter-event time $t_i - t_{i-1}$ is sampled i.i.d.\ from the same distribution with the cumulative hazard function $\Phi : \R_+ \to \R_+$.
The compensator of an RP is $\Lambda(t | \History_t) = \Phi(t - t_{i}) + \sum_{j=1}^i \Phi(t_{j} - t_{j-1})$, where $t_i$ is the last event before $t$.
The triangular map of an RP can be represented as a composition $\mapF = \mC \circ \mapPh \circ \mD$, where $\mD \in \R^{N \times N}$ is the pairwise difference matrix, $\mC \equiv \mD^{-1} \in \R^{N \times N}$ is the cumulative sum matrix, and $\mapPh$ applies $\Phi$ elementwise.

\textbf{Modulated renewal process (MRP)} \citep{cox1972statistical} 
generalizes both inhomogeneous Poisson and renewal processes.
The cumulative intensity is 
$\Lambda(t | \History_t) = \Phi(\Lambda(t) - \Lambda(t_{i})) + \sum_{j=1}^i \Phi(\Lambda(t_{j}) - \Lambda(t_{j-1}))$.
Again, we can represent the triangular map of an MRP as a composition, $\mapF = \mC \circ \mapPh \circ \mD \circ \mapLa$.

All three above models permit fast density evaluation and sampling.
Since $\mapPh$ and $\mapLa$ (as well as their inverses $\mapPh^{-1}$ and $\mapLa^{-1}$) are elementwise transformations, they can obviously be applied in $\gO(N)$ parallel operations.
Same holds for multiplication by the matrix $\mD$, as it is bidiagonal.
Finally, the cumulative sum defined by $\mC$ can also be computed in parallel in $\gO(N)$ \citep{blelloch1990prefix}.
Therefore, by reformulating IPP, RP and MRP using triangular maps, we can satisfy our efficiency requirements.

\textbf{Parametrization for $\Phi$ and $\Lambda$} must satisfy several conditions.
First, to define a valid TPP, $\Phi$ and $\Lambda$ have to be positive, strictly increasing and differentiable. 
Next, both functions, their derivatives (for density computation) and inverses (for sampling) must be computable in closed form to meet the efficiency requirements.
Lastly, we want both functions to be highly flexible.
Constructing such functions is not trivial.
While IPP, RP and MRP are established models, none of their existing parametrizations satisfy all the above conditions simultaneously.
Luckily, the same properties are necessary when designing normalizing flows~\citep{papamakarios2019normalizing}. 
Recently, Durkan et al.\ \citep{durkan2019neural} used rational quadratic splines (RQS) to define functions that satisfy our requirements.
We propose to use RQS to define $\Phi$ and $\Lambda$ for (M)RP and IPP.
This parametrization is flexible, while also allowing efficient density evaluation and sampling --- something that existing approaches are unable to provide (see \Secref{sec:related-work}).

\subsection{Defining more flexible triangular maps}
\label{sec:tri-tpp-triangular}
Even though the splines can make the functions $\Phi$ and $\Lambda$ arbitrarily flexible, the overall expressiveness of MRP is still limited.
Its conditional intensity $\sintensity(t)$ depends only on the global time and the time since the last event. 
This means, MRP cannot capture, e.g., self-exciting \citep{hawkes1971spectra} or self-correcting \citep{isham1979stochastic} behavior.
We will now construct a model that is more flexible without sacrificing the efficiency.

\begin{figure}[t]
    \centering
    \resizebox{0.99\textwidth}{!}{\input{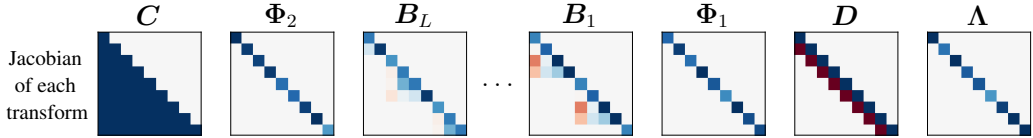}}
    \caption{\name defines an expressive map $\mapF$ as a composition of easy-to-invert transformations.}
    \label{fig:map-composition}
\end{figure}

The efficiency of the MRP stems from the fact that the respective triangular map $\mapF$ is defined as a composition of easy-to-invert transformations.
More specifically, we are combining \emph{learnable} element-wise nonlinear transformations $\mapPh$ and $\mapLa$ with \emph{fixed} lower-triangular matrices $\mD$ and $\mC$.
We can make the map $\mapF$ more expressive by adding \emph{learnable} lower-triangular matrices into the composition.
Using full $N \times N$ lower triangular matrices would be inefficient (multiplication and inversion are $\gO(N^2)$), and also would not work for variable-length sequences (i.e., arbitrary values of $N$).
Instead, we define block-diagonal matrices $\mB_l$, where each block is a repeated $H \times H$ lower-triangular matrix with strictly positive diagonal entries.
Computing $\mB_l^{-1}$ takes $\gO(H^2)$, and multiplication by $\mB_l$ or $\mB_l^{-1}$ can be done in $\gO(NH)$ in parallel.
We stack $L$ such matrices $\mB_l$ and define the triangular map
$\mapF = \mC \circ \mapPh_2 \circ \mB_L \circ \cdots \circ \mB_1 \circ \mapPh_1 \circ \mD \circ \mapLa$.
The blocks in every other layer are shifted by an offset $H/2$ to let the model capture long-range dependencies.
Note that now we use two element-wise learnable splines $\mapPh_1$ and $\mapPh_2$ before and after the block-diagonal layers. 
\Figref{fig:map-composition} visualizes the overall sequence of maps and the Jacobians of each transformation.
We name the temporal point process densities defined by the triangular map $\mapF$ as \name.

Both the forward map $\mapF$ and its inverse $\mapF^{-1}$ can be evaluated in parallel in linear time, making \name efficient for density computation and sampling.
Our insight that TPP densities can be represented by increasing triangular maps was crucial for arriving at this result.
Alternative representations of \name, e.g., in terms of the compensator $\CumulativeIntensity$ or the conditional intensity $\sintensity$, are cumbersome and do not emphasize the parallelism of the model.
\name and our parametrizations of IPP, RP, MRP can be efficiently implemented on GPU to handle batches of variable-length sequences (\Appref{app:impl-details}).

\section{Differentiable sampling-based losses for temporal point processes}
\label{sec:diff}

Fast parallel sampling allows us to efficiently answer prediction queries such as "How many events are expected to happen in the next hour given the history?".
More importantly, it enables us to efficiently train TPP models using objective functions of the form $\E_{p}[g(\vt)]$.
This includes using $p(\vt)$ to specify the policy in reinforcement learning \citep{upadhyay2018deep}, to impute missing data during training \citep{shchur2020intensity} or to define an approximate posterior in variational inference (\Secref{sec:diff-mjp}).
In all but trivial cases the expression $\E_{p}[g(\vt)]$ has no closed form, so we need to estimate its gradients w.r.t.\ the parameters of $p(\vt)$ using Monte Carlo (MC).
Recall that we can sample from $p(\vt)$ by applying the map $\mapF^{-1}$ to $\vz$ drawn from an HPP with unit rate.
This enables the so-called reparametrization trick \citep{mohamed2019monte}.
Unfortunately, this is not enough.
Sampling-based losses for TPPs are in general not differentiable.
This is a property of the loss functions that is independent of the parametrization of $p(\vt)$ or the sampling method.
In the following, we provide a simple example and a solution to this problem.

\subsection{Entropy maximization}
\label{sec:diff-entropy}
Consider the problem of maximizing the entropy of a TPP.
An entropy penalty can be used as a regularizer during density estimation \citep{grandvalet2006entropy} 
or as a part of the ELBO in variational inference.
Let $p_\lambda(\vt)$ be a homogeneous Poisson process on $[0, T]$ with rate $\lambda > 0$.
It is known that the entropy is maximized when $\lambda = 1$ \citep{baccelli2016entropy}, 
but for sake of example assume that we want to learn $\lambda$ that maximizes the entropy 
$-\E_{p}[\log p_\lambda(\vt)]$ with gradient ascent.
We sample from $p_\lambda(\vt)$ by drawing a sequence $\vz = (z_1, z_2, ...)$ from a HPP with unit rate
and applying the inverse map $\vt = \mapF^{-1}_\lambda(\vz) = \frac{1}{\lambda} \vz$ (\Figref{fig:sampling}).
We obtain an MC estimate of the entropy using a single such sample $\vt = (t_1, t_2, ...)$ as
\begin{align}
    \label{eq:nondiff}
    -\E_{p} [\log p_\lambda(\vt)] \approx \lambda T - \sum_{i=1}^\infty \vone(t_i \le T) \log \lambda
    = \lambda T - \sum_{i=1}^\infty \vone\left(\frac{1}{\lambda} z_i \le T\right) \log \lambda 
\end{align}
Here, the indicator function $\vone(\cdot)$ discards all the events $t_i > T$.
We can see that for any sample $\vz$ the right-hand side of \Eqref{eq:nondiff} is not continuous w.r.t.\ $\lambda$ at points $\lambda = \frac{1}{T}z_i$.
At such points, decreasing $\lambda$ by an infinitesimal amount will "push" the sample $t_i = \frac{1}{\lambda}z_i$ outside the $[0, T]$ interval, thus increasing $\log p_\lambda(\vt)$ by a constant $\log \lambda$.
We plot the right-hand side of \Eqref{eq:nondiff} as a function of $\lambda$ in \Figref{fig:nondiff-relaxation}, estimated with 5 MC samples.
Clearly, such function cannot be optimized with gradient ascent.
Increasing the number of MC samples almost surely adds more points of discontinuity and does not fix the problem. 
In general, non-differentiability arises when estimating expectations of a function $g(\vt)$ that depends on the events $t_i$ inside $[0, T]$.
For any TPP density $p(\vt)$, the discontinuities occur at the parameter values that map the HPP realizations $z_i$ exactly to the interval boundary $T$.

\begin{figure}
    \centering
    \begin{minipage}{0.31\textwidth}
        \centering
        \resizebox{!}{80pt}{\input{figures/relax_juxt.pgf}}
        \vspace{-3mm}
        \caption{Monte Carlo estimate of the entropy.}%
        \label{fig:nondiff-relaxation}
    \end{minipage}
    \hfill
    \begin{minipage}{0.33\textwidth}
        \centering
        \resizebox{!}{80pt}{\input{figures/convergence.pgf}}
        \vspace{-3mm}
        \caption{Maximizing the entropy with different values of $\gamma$.}
        \label{fig:nondiff-convergence}
    \end{minipage}
    \hfill
    \begin{minipage}{0.30\textwidth}
	\centering
	\resizebox{!}{80pt}{\input{figures/mmpp.pgf}}
	\vspace{-3mm}
	\caption{Markov modulated Poisson process with 2 states.}
	\label{fig:mmpp}
    \end{minipage}
    \vspace{0mm}
\end{figure}

\textbf{Relaxation.}
We obtain a differentiable approximation to \Eqref{eq:nondiff} by relaxing the indicator functions as
$\vone(t_i \le T) \approx \sigmoid_\gamma(T - t_i)$, where $\sigmoid_\gamma(x) = 1/(1 + \exp(-x/\gamma))$ is the sigmoid function with a temperature parameter $\gamma > 0$.
Decreasing the temperature $\gamma$ makes the approximation more accurate, but complicates optimization, similarly to the Gumbel-softmax trick \citep{jang2017categorical}.
\Figref{fig:nondiff-convergence} shows convergence plots for different values of $\gamma$.
Our relaxation applies to MC estimation of any function $g(\vt)$ that can be expressed in terms of the indicator functions.
This method also enables differentiable sampling with reparametrization from a Poisson distribution, which might be of independent interest.

\subsection{Variational inference for Markov jump processes}
\label{sec:diff-mjp}
Combining fast sampling (\Secref{sec:tri-tpp}) with the differentiable relaxation opens new applications for TPPs.
As an example, we design a variational inference scheme for Markov jump processes.

\textbf{Background.} A Markov jump process (MJP) $\{s(t)\}_{t \ge 0}$ is a piecewise-constant stochastic process on $\R_+$.
At any time $t$, the process occupies a discrete state $s(t) \in \{1, ..., K\}$.
The times when the state changes are called jumps.
A trajectory of an MJP on an interval $[0, T]$ with $N$ jumps can be represented by a
tuple $(\vt, \vs)$ of jump times $\vt = (t_1, ..., t_N)$ and the visited states $\vs = (s_1, ..., s_{N+1})$.
Note that $N$ may vary for different trajectories.
The prior over the trajectories $p(\vt, \vs| \vpi, \mA)$
of an MJP is governed by an initial state distribution $\vpi$ and a $K \times K$ generator matrix $\mA$ (see \Appref{app:vi-mmpp}).

MJPs are commonly used to model the unobserved (latent) state of a system.
In a latent MJP, the state $s(t)$ influences the behavior of the system and indirectly manifests itself via some observations $\obs$.
For concreteness, we consider the Markov-modulated Poisson process (MMPP) \citep{fischer1993markov}.
In an MMPP, each of the $K$ states of the MJP has an associated observation intensity $\lambda_k$.
An MMPP is an inhomogeneous Poisson process where the intensity depends on the current MJP state as $\intensity(t) = \lambda_{s(t)}$.
For instance, a 2-state MMPP can model the behavior of a social network user, who switches between an "active" (posting a lot) and "inactive" (working or sleeping) states (\Figref{fig:mmpp}).
Given the observations $\obs$, we might be interested in inferring the trajectory $(\vt, \vs)$, the model parameters $\vtheta = \{\vpi, \mA, \vlambda\}$, or both.

\textbf{Variational inference.}
The posterior distribution $p(\vt, \vs | \obs, \vtheta)$ of MMPP is intractable, so we approximate it with a variational distribution $q(\vt, \vs) = q(\vt) q(\vs | \vt)$.
Note that this is \emph{not} a mean-field approximation used in other works \citep{zhang2017collapsed}.
We model the distribution over the jump times $q(\vt)$ with \name (\Secref{sec:tri-tpp-triangular}).
We find the best approximate posterior by maximizing the ELBO \citep{zhang2018advances}
\begin{align}
    \label{eq:elbo}
    \max_{q(\vt)} \; \max_{q(\vs | \vt)}\; \E_{q(\vt)}
    \left[ \E_{q(\vs | \vt)} 
    \left[
        \log p(\obs | \vt, \vs, \vtheta) + \log p(\vt, \vs | \vtheta) - \log q(\vt, \vs)
    \right]
    \right]
\end{align}
Given jump times $\vt$, the true posterior over the states $p(\vs | \vt, \obs, \vtheta)$ is just the posterior of a discrete hidden Markov model (HMM).
This means that we only need to model $q(\vt)$; the optimal $q^\star(\vs | \vt)$, i.e.
\begin{align}
    \label{eq:variational-state}
    q^\star(\vs | \vt) &= \argmax_{q(\vs | \vt)} \E_{q(\vs | \vt)}\left[\log p(\obs | \vt, \vs, \vtheta) + \log p(\vt, \vs| \vtheta) - \log q(\vs | \vt)\right]
    = p(\vs | \vt, \obs, \vtheta)
\end{align}
can be found by doing inference in an HMM --- doable efficiently via the forward-backward algorithm \citep{wang2013collapsed}. 
The inner expectation w.r.t.\ $q(\vs|\vt)$ in \Eqref{eq:elbo} can be computed analytically. 
We approximate the expectation w.r.t.\ $q(\vt)$ with Monte Carlo.
Since all terms of \Eqref{eq:elbo} are not differentiable, we apply our relaxation from \Secref{sec:diff-entropy}.
We provide a full derivation of the ELBO and the implementation details in \Appref{app:vi-elbo}.

The proposed framework is not limited to approximating the posterior over the trajectories.
With small modifications (\Appref{app:vi-param-estimation}), we can simultaneously learn the parameters $\vtheta$, either obtaining a point estimate $\vtheta^\star$ or a full approximate posterior $q(\vtheta)$.
Our variational inference scheme can also be extended to other continuous-time discrete-state models, 
such as semi-Markov processes~\citep{feller1964semi}.

\section{Related work}
\label{sec:related-work}
\textbf{Triangular maps} \citep{jaini2019sum} can be seen as a generalization of autoregressive normalizing flows \citep{germain2015made,kingma2016improved,papamakarios2019normalizing}.
Existing normalizing flow models  are either limited to fixed-dimensional data \citep{dinh2016density,papamakarios2017masked} or are inherently sequential \citep{oord2016wavenet,van2016conditional}.
Our model proposed in \Secref{sec:tri-tpp-triangular} can handle variable-length inputs, and allows for both
$\mapF$ and $\mapF^{-1}$ to be evaluated efficiently in parallel.

\textbf{Sampling from TPPs.}
Inverse method for sampling from inhomogeneous Poisson processes can be dated back to \c{C}inlar \citep{cinlar1975introduction}.
However, traditional inversion methods for IPPs are different from our approach (\Secref{sec:tri-tpp}).
First, they are typically sequential.
Second, existing methods either use extremely basic compensators $\Lambda(t)$, such as $\lambda t$ or $e^{\alpha t}$, or require numerical inversion \citep{pasupathy2010generating}.
As an alternative to inversion, thinning approaches \citep{lewis1979simulation} became the dominant paradigm for generating IPPs, and TPPs in general.
Still, sampling via thinning has a number of disadvantages.
Thinning requires a piecewise-constant upper bound on $\lambda(t)$, which might not always be easy to find.
If the bound is not tight, a large fraction of samples will be rejected.
Moreover, thinning is not differentiable, doesn't permit reparametrization, and is hard to express in terms of parallel operations on tensors~\citep{turkmen2019fastpoint}.
Our inversion-based sampling addresses all the above limitations.
It's also possible to generate an IPP by first drawing $N \sim \operatorname{Poisson}(\Lambda(T))$ and then sampling $N$ points $t_i$ i.i.d.\ from a density $p(t) = \lambda(t) / \Lambda(T)$
\citep{cox1966statistical}.
Unlike inversion, this method is only applicable to Poisson processes.
Also, the operation of sampling $N$ is not differentiable, which limits the utility of this approach.

\textbf{Inhomogeneous Poisson processes} are commonly defined by specifying the intensity function $\lambda(t)$ via a latent Gaussian process \citep{cox1955some}.
Such models are flexible, but highly intractable.
It’s possible to devise approximations by, e.g., bounding the intensity function \citep{adams2009tractable,donner2018efficient}.
Our spline parametrization of IPP compares favorably to the above models: it is also highly flexible, has a tractable likelihood and places no restrictions on the intensity. 
Importantly, it is much easier to implement and train. 
If uncertainty is of interest, we can perform approximate Bayesian inference on the spline coefficients \citep{zhang2018advances}.
Recently, Morgan et al. \citep{morgan2019spline} used splines to model the intensity function of IPPs. 
Since $\Lambda^{-1}$ cannot be computed analytically for their model, sampling via thinning is the only available option.

\textbf{Modulated renewal processes}
have been known for a long time \citep{cox1972statistical,berman1981inhomogeneous},
but haven't become as popular as IPPs among practitioners.
This is not surprising, since inference and sampling in MRPs are even more challenging than in Cox processes \citep{rao2011gaussian,lasko2014efficient}.
Our proposed parametrization addresses the shortcomings of existing approaches and makes MRPs straightforward to apply in practice.

\textbf{Neural TPPs.}
Du et al.\ \citep{du2016rmtpp} proposed a TPP model based on a recurrent neural network.
Follow-up works improved the flexibility of RNN-based TPPs by e.g.\ changing the RNN architecture \citep{mei2017neural}, using more expressive conditional hazard functions \citep{omi2019fully,bilovs2019uncertainty} or modeling the inter-event time distribution with normalizing flows \citep{shchur2020intensity}.
All the above models are inherently sequential and therefore inefficient for sampling (\Secref{sec:exp-scale}).
Recently, Turkmen et al.\ \citep{turkmen2019fastpoint} proposed to speed up RNN-based \emph{marked} TPPs by discretizing the interval $[0, T]$ into a regular grid.
Samples within each grid cell can be produced in parallel for each mark, but the cells themselves still must be processed sequentially.

\textbf{Latent space models.}
TPPs governed by latent Markov dynamics have intractable likelihoods that require approximations \citep{hirt2019scalable,wu2019markov}.
For MJPs, the state-of-the-art approach is the Gibbs sampler by Rao \& Teh \citep{rao2013fast}.
It allows to exactly sample from the posterior $p(\vt, \vs | \obs, \vtheta)$, but is known to converge  slowly if the parameters $\vtheta$ are to be learned as well \citep{zhang2019efficient}.
Existing variational inference approaches for MJPs can only learn a fixed time discretization \cite{zhang2017collapsed} or estimate the marginal statistics of the posterior \cite{opper2008variational, wildner2019moment}.
In contrast, our method (\Secref{sec:diff-mjp}) produces a full distribution over the jump times.

\section{Experiments}
\label{sec:exp}

\subsection{Scalability}
\label{sec:exp-scale}
\textbf{Setup.}
The key feature of \name is its ability to compute likelihood and generate samples in parallel, which is impossible for RNN-based models.
We quantify this difference by measuring the runtime of the two models.
We implemented \name and RNN models in PyTorch \citep{PyTorch}.
The architecture of the RNN model is nearly identical to the ones used in \citep{du2016rmtpp,omi2019fully,shchur2020intensity},
except that the cumulative conditional hazard function is parametrized with a spline \citep{durkan2019neural} to enable closed-form sampling.
\Appref{app:setup} contains the details for this and other experiments.
We measure the runtime of (a) computing the log-likelihood (and backpropagate the gradients) for a batch of 100 sequences of varying lengths and (b) sample sequences of the same sizes.
We used a machine with an Intel Xeon E5-2630 v4 @ 2.20 GHz CPU, 256GB RAM and an Nvidia GTX1080Ti GPU.
The results are averaged over 100 runs.

\begin{wrapfigure}{r}{0.30\textwidth}
    \centering
    \vspace{-5mm}
    \resizebox{0.30\textwidth}{!}{\input{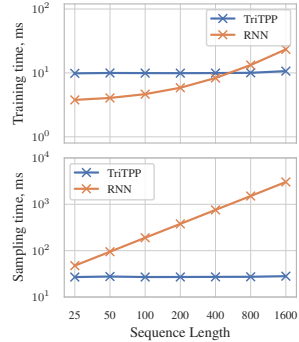}}
    \vspace{-3mm}
    \caption{Scalability analysis. Standard devs. are below 1ms.}
    \label{fig:scalability}
\end{wrapfigure}
\textbf{Results.}
\Figref{fig:scalability} shows the runtimes for varying sequence lengths.
Training is rather fast for both models, on average taking 1-10ms per iteration.
RNN is slightly faster for short sequences, but is outperformed by \name on sequences with more than 400 events.
Note that during training we used a highly optimized RNN implementation based on custom CUDA kernels (since all the event times $t_i$ are already known).
In contrast, \name is implemented using generic PyTorch operations.
When it comes to sampling, we notice a massive gap in performance between \name and the RNN model.
This happens because RNN-based TPPs are defined autoregressively and can only produce samples $t_i$ one by one:
to obtain $p(t_i | t_1, ..., t_{i-1})$ we must know all the past events.
Recently proposed transformer TPPs \citep{zhang2020self,zuo2020transformer} are defined in a similar autoregressive way, so they are likely to be as slow for sampling as RNNs.
\name generates all the events in a sequence in parallel, which makes it more than 100 times faster than the recurrent model for longer sequences.

\subsection{Density estimation}
\label{sec:exp-prediction}
\textbf{Setup.}
A fast TPP model is of little use if it cannot accurately learn the data distribution.
The main goal of this experiment is to establish whether \name can match the flexibility of RNN-based TPPs.
As baselines, we use the IPP, RP and MRP models from \Secref{sec:tri-tpp-nonar} and Hawkes process \citep{tick}.

\textbf{Datasets.}
We use 6 synthetic datasets from Omi et al. \citep{omi2019fully}:
Hawkes1\&2 \citep{hawkes1971spectra}, self-correcting (SC) \citep{isham1979stochastic}, inhomogeneous Poisson (IPP), renewal (RP) and modulated renewal (MRP) processes.
Note that the data generators for IPP, RP and MRP by Omi et al.\ are \emph{not} parametrized using splines, so these datasets are not guaranteed to be fitted perfectly by our models.
We also consider 7 real-world datasets: 
PUBG (online gaming), Reddit-Comments, Reddit-Submissions (online discussions), Taxi (customer pickups), Twitter (tweets) and Yelp1\&2 (check-in times).
See \Appref{app:datasets} for more details.

\textbf{Metrics.}
The standard metric for comparing generative models, including TPPs, is negative log-likelihood (NLL) on a hold-out set \citep{turkmen2019fastpoint,omi2019fully,shchur2020intensity}.
We partitioned the sequences in each dataset into train/validation/test sequences (60\%/20\%/20\%).
We trained the models by minimizing the NLL of the train set using Adam \citep{kingma2014adam}.
We tuned the following hyperparameters: $L_2$ 
regularization $\{0, 10^{-5}, 10^{-4}, 10^{-3}\}$, number of spline knots $\{10, 20, 50\}$, learning rate $\{10^{-3}, 10^{-2}\}$,
hidden size $\{32, 64\}$ for RNN, 
number of blocks $\{2, 4\}$ and block size $\{8, 16\}$ for \name.
We used the validaiton set for hyperparameter tuning, early stopping and model development.
We computed the results for the test set only once before including them in the paper.
All results are averaged over 5 runs.

While NLL is a popular metric, it has known failure modes \citep{theis2015note}.
For this reason, we additionally computed maximum mean discrepancy (MMD) \citep{gretton2012kernel} between the test sets and the samples drawn from each model after training.
To measure similarity between two realizations $\vt$ and $\vt^\prime$, we use a Gaussian kernel 
$k(\vt, \vt^\prime) = \exp(-d(\vt, \vt^\prime) / 2\sigma^2)$,
where $d(\vt, \vt^\prime)$ is the "counting measure" distance from 
\citep[Equation 3]{xiao2017wasserstein}.
For completeness, we provide the definitions in \Appref{app:setup-density}.
MMD quantifies the dissimilarity between the true data distribution $p^\star(\vt)$ and the learned density $p(\vt)$ --- lower is better.

\begin{table}
    \caption{Average test set NLL on synthetic and real-world datasets (lower is better). Best NLL in \textbf{bold}, second best \underline{underlined}. 
    Results with standard deviations can be found in \Appref{app:additional-density}.}
    \label{tab:nll-results}
    \vspace{2mm}
    \centering
    \resizebox{\columnwidth}{!}{% \begin{tabular}{lrrrrrr}
% {} &   Taxi &  Twitter &  Yelp1 &  Yelp2 &  R-Comments &  R-Submissions \\
% \midrule
% Inhom. Poisson       & -0.677 & 1.601 & 0.616 & -0.054 & -1.586 & -4.077 \\
% Renewal              & -0.576 & 1.196 & 0.675 &  0.016 & -2.084 & -3.918 \\
% Modulated Renewal    & -0.677 & 1.231 & 0.610 &  0.102 & -2.134 & -4.096 \\
% Hawkes               & -0.663 & 1.081 & 0.672 &  0.075 & -2.401 & -4.254 \\
% Autoregressive       & -0.663 & 1.081 & 0.672 &  0.075 & -2.401 & -4.254 \\
% \name                & -0.641 & 1.067 & 0.666 &  0.061 & -2.362 & -4.171 \\
% \end{tabular}

\begin{tabular}{l|rrrrrr|rrrrrrr}
{} &  Hawkes1 &  Hawkes2 &    SC &   IPP &   MRP &    RP &  PUBG &  Reddit-C &  Reddit-S &  Taxi &  Twitter &  Yelp1 &  Yelp2 \\
\midrule
IPP    &         1.06 &            1.03 &         1.00 & \textbf{0.71} &         0.70 &         0.89 &        -0.06 &          -1.59 &          -4.08 & \textbf{-0.68} &           1.60 &   \second{0.62} &          -0.05 \\
RP     &         0.65 &            0.08 &         0.94 &         0.85 &         0.68 & \textbf{0.24} &         0.12 &          -2.08 &          -4.00 &         -0.58 &           1.20 &           0.67 &          -0.02 \\
MRP    &         0.65 &            0.07 &         0.93 & \textbf{0.71} &         0.36 &         0.25 &        -0.83 &          -2.13 &          -4.38 & \textbf{-0.68} &           1.23 & \textbf{0.61} & \textbf{-0.10} \\
Hawkes &\textbf{0.51} &            0.06 &         1.00 &         0.86 &         0.98 &         0.39 &         0.11 &  \textbf{-2.40} &          -4.19 &         -0.64 & \textbf{1.04} &           0.69 &           0.01 \\
RNN    & \second{0.52} &  \textbf{-0.03} & \textbf{0.79} &         0.73 & \second{0.37} & \textbf{0.24} &\second{-1.96} &  \textbf{-2.40} &  \textbf{-4.89} &         -0.66 &           1.08 &           0.67 &          -0.08 \\
TriTPP &         0.56 &   \second{0.00} & \second{0.83} & \textbf{0.71} & \textbf{0.35} & \textbf{0.24} &\textbf{-2.41} &          -2.36 &  \second{-4.49} &         -0.67 &   \second{1.06} &           0.64 &  \second{-0.09}\\
\end{tabular}
}
\end{table}
\begin{table}
    \caption{MMD between the hold-out test set and the generated samples (lower is better). 
    }
    \label{tab:mmd-results}
    \vspace{2mm}
    \centering
    \resizebox{\columnwidth}{!}{\begin{tabular}{l|rrrrrr|rrrrrrr}
{} &  Hawkes1 &  Hawkes2 &    SC &   IPP &   MRP &    RP &  PUBG &  Reddit-C &  Reddit-S &  Taxi &  Twitter &  Yelp1 &  Yelp2 \\
\midrule
% IPP    &          0.08&         0.09&         0.59&\textbf{0.02}&         0.15&         0.07&\textbf{0.02}&         0.10&         0.18&         0.10&         0.16&\second{0.14}&\second{0.16} \\
% RP     &          0.06&         0.06&         1.12&         0.35&         1.24&\textbf{0.01}&         0.49&         0.06&         0.16&         0.58&         0.14&         0.15&         0.23 \\
% MRP    &          0.05&         0.06&         0.51&\textbf{0.02}&\second{0.11}&\textbf{0.01}&\second{0.12}&         0.08&         0.18&\second{0.09}&         0.13&\second{0.14}&         0.17 \\
% Hawkes & \second{0.02}&\second{0.04}&         0.60&         0.36&         0.64&         0.05&         0.17&\textbf{0.04}&         0.32&         0.20&         0.21&         0.20&         0.32 \\
% RNN    & \textbf{0.01}&\textbf{0.01}&\textbf{0.21}&         0.09&         0.17&\textbf{0.01}&         0.25&\textbf{0.04}&\textbf{0.08}&         0.13&\textbf{0.08}&         0.19&         0.19 \\
% TriTPP &          0.03&\second{0.04}&\second{0.25}&\textbf{0.02}&\textbf{0.08}&\textbf{0.01}&         0.17&         0.07&\second{0.14}&\textbf{0.08}&\second{0.09}&\textbf{0.12}&\textbf{0.14} \\

IPP    &          0.08&         0.09&         0.58&\textbf{0.02}&         0.15&         0.07&\textbf{0.01}&         0.10&         0.21&         0.10&         0.16&         0.15&\second{0.16} \\
RP     &          0.06&         0.06&         1.13&         0.34&         1.24&\textbf{0.01}&         0.46&         0.07&         0.18&         0.57&         0.14&         0.16&         0.23 \\
MRP    &          0.05&         0.06&         0.50&\textbf{0.02}&\second{0.11}&         0.02&\second{0.12}&         0.09&         0.20&\second{0.09}&         0.13&\second{0.13}&\second{0.16} \\
Hawkes & \second{0.02}&         0.04&         0.58&         0.36&         0.65&         0.05&         0.16&\textbf{0.04}&         0.35&         0.20&         0.20&         0.20&         0.32 \\
RNN    & \textbf{0.01}&\textbf{0.02}&\textbf{0.19}&         0.09&         0.17&\textbf{0.01}&         0.23&\textbf{0.04}&\textbf{0.09}&         0.13&\textbf{0.08}&         0.19&         0.18 \\
TriTPP &          0.03&\second{0.03}&\second{0.23}&\textbf{0.02}&\textbf{0.08}&\textbf{0.01}&         0.16&         0.07&\second{0.16}&\textbf{0.08}&\textbf{0.08}&\textbf{0.12}&\textbf{0.14} \\

\end{tabular}
}
\end{table}

\textbf{Results.}
Table \ref{tab:nll-results} shows the test set NLLs for all models and datasets.
We can see that the RNN model achieves excellent scores and outperforms the simpler baselines, which is consistent with earlier findings \citep{du2016rmtpp}.
\name is the only method that is competitive with the RNN --- our method is within 0.05 nats of the best score on 11 out of 13 datasets.
\name consistently beats MRP, RP and IPP, which confirms that learnable block-diagonal transformations improve the flexibility of the model.
The gap get larger on the datasets such as Hawkes, SC, PUBG and Twitter, where the inability of MRP to learn self-exciting and self-correcting behavior is especially detrimental.
While Hawkes process is able to achieve good scores on datasets with "bursty" event occurrences (Reddit, Twitter), it is unable to adequately model other types of behavior (SC, MRP, PUBG).

Table \ref{tab:mmd-results} reports the MMD scores.
The results are consistent with the previous experiment: models with lower NLL typically obtain lower MMD.
One exception is the Hawkes process that achieves low NLL but high MMD on Taxi and Twitter.
\name again consistently demonstrates excellent performance.
Note that MMD was computed using the test sequences that were unseen during training.
This means that \name models the data distribution better than other methods, and does not just simply overfit the training set.
In \Appref{app:additional-density}, we provide additional experiments for quantifying the quality of the distributions learned by different models.
Overall, we conclude that \name is flexible and able to model complex densities, in addition to being significantly more efficient than RNN-based TPPs.

\subsection{Variational inference}
\label{sec:exp-vi}
\begin{wrapfigure}{r}{0.4\textwidth}
    \vspace{-8mm}
    \includegraphics[width=0.4\columnwidth]{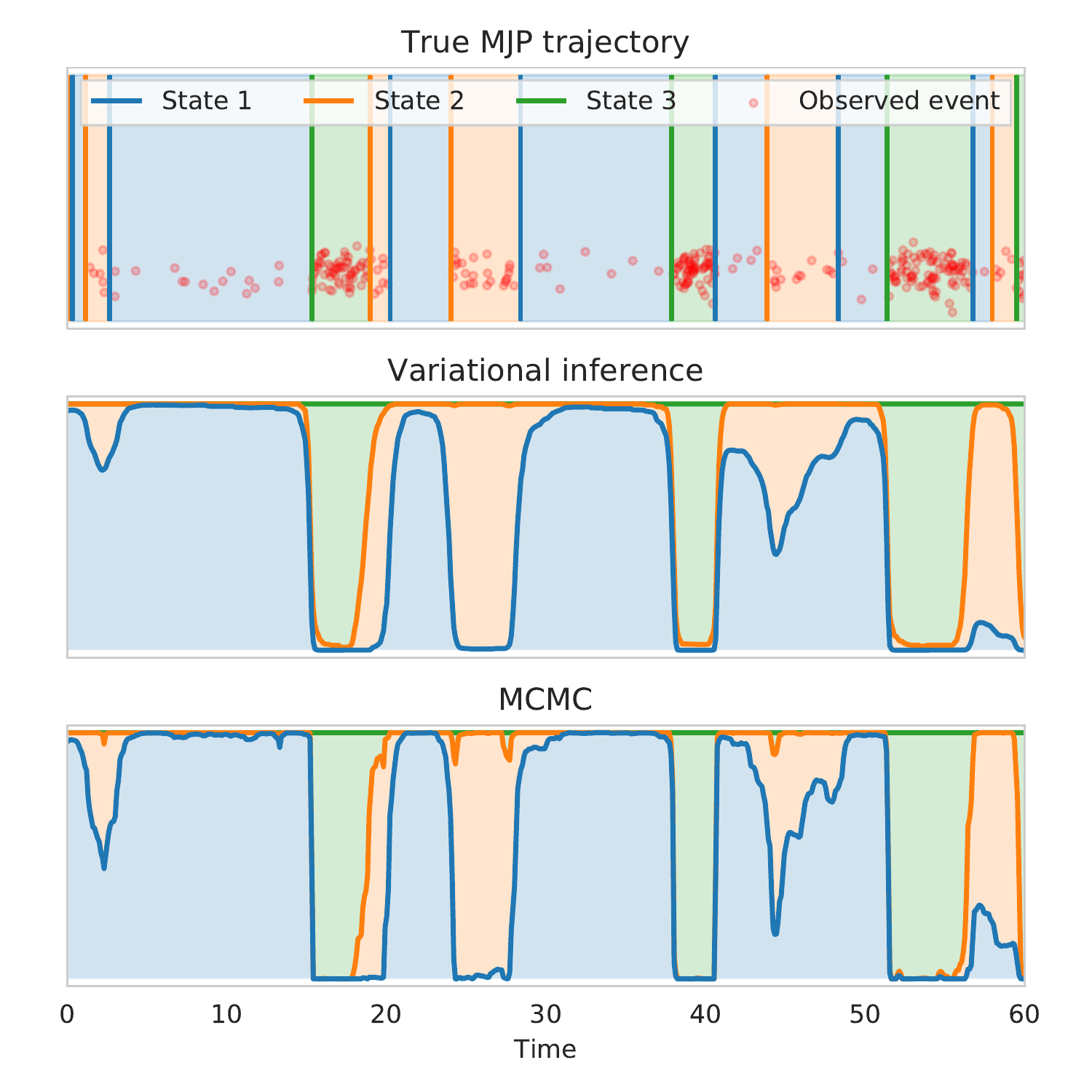} 
    \vspace{-7mm}
    \caption{Posterior distributions over the latent trajectory of an MMPP learned using our VI approach \& MCMC.}
    \vspace{-9mm}
    \label{fig:vi-posterior}
\end{wrapfigure}
\textbf{Setup.}
We apply our variational inference method (\Secref{sec:diff-mjp}) for learning the posterior distribution over the latent trajectories of an MMPP.
We simulate an MMPP with $K=3$ latent states.
As a baseline, we use the state-of-the-art MCMC sampler by Rao \& Teh \citep{rao2013fast}.

\textbf{Results.}
\Figref{fig:vi-posterior} shows the true latent MJP trajectory, as well as the marginal posterior probabilities learned by our method and the MCMC sampler of Rao \& Teh.
We can see that \name accurately recovers the true posterior distribution over the trajectories.
The two components that enable our new variational inference approach are our efficient parallel sampling algorithm for \name (\Secref{sec:tri-tpp}) and the differential relaxation (\Secref{sec:diff}).
\Appref{app:additional-vi} contains an additional experiment on real-world data, where we both learn the parameters $\vtheta$ and infer the posterior over the trajectories.

\section{Future work \& conclusions}
\textbf{Future work \& limitations.}
We parametrized the nonlinear transformations of our TPP models with splines.
Making a spline more flexible requires increasing the number of knots, which increases the number of parameters and might lead to overfitting.
New deep \emph{analytically} \emph{invertible} functions will improve both our models, as well as normalizing flows in general.
Currently, \name is not applicable to marked TPPs \citep{rasmussen2011temporal}.
Extending our model to this setting is an important task for future work.

\textbf{Conclusions.}
We have shown that TPP densities can be represented with increasing triangular maps.
By directly parametrizing the respective transformations, we are able to construct TPP models, for which both density evaluation and sampling can be done efficiently in parallel.
Using the above framework, we defined \name --- a new class of flexible probability distributions over variable-length sequences.
In addition to being highly efficient thanks to its parallelism,
\name shows excellent performance on density estimation, as shown by our experiments.
High flexibility and efficiency of \name allow it to be used as a plug-and-play component of other machine learning models.

\section*{Broader impact}
Existing works have applied TPPs and MJPs for analyzing electronic health records \citep{alaa2018hidden,cao2016continuous}, detecting anomalies in network traffic \citep{ihler2007learning,yan2007anomaly} and modeling user behavior on online platforms \citep{rodriguez2011uncovering,mavroforakis2017modeling}. 
Thanks to fast sampling, our model can be used for solving new prediction tasks on such data, and the overall improved scalability allows practitioners to work with larger datasets.
We do not find any of the above use cases ethically questionable, though, general precautions must be implemented when handling sensitive personal data.
Since our model exploits fast parallel computations, has fewer parameters and converges in fewer iterations, it is likely to be more energy-efficient compared to RNN-based TPPs.
However, we haven't performed experiments analyzing this specific aspect of our model.

\section*{Acknowledgments}
This research was supported by the German Federal Ministry of Education and Research (BMBF),
grant no. 01IS18036B, the Software Campus Project Deep-RENT and by the BMW AG. The authors of this work take
full responsibilities for its content.

\bibliography{references}
\bibliographystyle{unsrt}

\appendix
\newpage
\section{Abbreviations and notation}
\label{app:notation}
Abbreviations:
\begin{itemize}
    \item RNN --- recurrent neural network
    \item TPP --- temporal point process
    \item HPP --- homogeneous Poisson process (compensator is $\CumulativeIntensity(t) = \lambda t$ for some $\lambda > 0$)
    \item IPP --- inhomogeneous Poisson process %
    \item RP --- renewal process %
    \item MRP --- modulated renewal process %
    \item MJP -- Markov jump process %
    \item MMPP --- Markov modulated Poisson process %
    \item MC --- Monte Carlo
    \item VI --- variational inference
\end{itemize}

\vspace{2\baselineskip}
\def\arraystretch{1.7}
\begin{table}[h]
    \centering
    \caption{Notation used throughout the paper.}
    \label{tab:notation}
    \begin{tabularx}{\columnwidth}{c|X}
        Notation & Description\\
        \midrule
        $\vt = (t_1, ..., t_N)$ & Variable-length realization of a TPP.\\ \hline
        $p(\vt)$ & Density of a point process, also called likelihood (\Eqref{eq:tpp-likelihood}).\\ \hline
        $\sintensity(t) = \lambda(t|t_1, ..., t_{i-1})$ &
        Conditional intensity at time $t$, where $t_{i-1}$ is the last event before $t$.\\ \hline
        \vtop{\hbox{\strut $\CumulativeIntensity(t) = \Lambda(t|t_{1}, ..., t_{i-1})$}\hbox{\strut $ \;\; \qquad = \int_0^T \sintensity(u) du$}}
          & Cumulative conditional intensity at time $t$, also known as the compensator.\\  \hline
        $\Lambda(t)$ & (Unconditional) cumulative intensity of a Poisson process. \\ \hline
        $\Phi(\tau)$ & Cumulative hazard function of a renewal process. \\ \hline
        $\mC$ & The $N \times N$ cumulative sum matrix,
        $C_{ij} = \begin{cases}
            1 & \text{ if } i \le j,\\
            0 & \text{ else.}
        \end{cases}$
        \\ \hline
        $\mD \equiv \mC^{-1}$ & The $N \times N$ difference matrix,
        $D_{ij} = \begin{cases}
            1 & \text{ if } i = j,\\
            -1 & \text{ if } i = j + 1,\\
            0 & \text{ else.}
        \end{cases}$\\ \hline
        $\mapF = (f_1, ..., f_N)$ & Increasing lower-triangular map that converts a realization $\vt$ of an arbitrary TPP with compensator $\CumulativeIntensity$ into a sample $\vz$ from an HPP with unit rate.\\ \hline
        $f_i(t_1, ..., t_i) = \Lambda(t_i | t_1, ..., t_{i-1})$ & Component function of $\mapF$.\\ \hline
        $\vone(x)$ & Indicator function, $\vone(x) = \begin{cases}
            1 & \text{ if } x \text{ is True},\\
            0 & \text{ else.}
        \end{cases}$\\ \hline
        $\gamma$ & Temperature parameter for the diff.\ relaxation (\Secref{sec:diff-entropy}).
    \end{tabularx}
\end{table}

\newpage

\section{Variational inference for Markov jump processes}
\label{app:vi}
\subsection{Generative model for MJP and MMPP}
\label{app:vi-mmpp}
\textbf{Markov jump process.}
We represent the trajectory of an MJP as a tuple $(\vt, \vs)$, where $\vt = (t_1, ..., t_N)$ are the (strictly increasing) jump times and $\vs = (s_1, ..., s_{N+1})$ is the sequence of visited states.
For convenience, we additionally set $t_0 = 0$ and $t_{N+1} = T$.

The distribution over the trajectories $(\vt, \vs)$ is defined by a $K \times K$ generator matrix $\mA$ and an initial state distribution $\vpi$.
Each entry $A_{kl} \in \R_+$ denotes the rate of transition from state $k$ to state $l$ of the the MJP.
Note that we use the formulation that permits self-jumps \citep{rao2012mcmc} (i.e., it may happen that $s_i = s_{i+1}$).
We can denote the total transition rate of state $s_i$ as $A_{s_i} = \sum_{k=1}^{K} A_{s_i k}$.
We can simulate an MJP trajectory using the following procedure
\begin{align}
    \begin{split}
    &s_1 \sim \Categorical(\vpi)\\
    &t_i - t_{i-1} =: \tau_i  \sim \Exponential\left(A_{s_i}\right)\\
    &s_{i+1} \sim \Categorical\left(\mA_{s_i :} / A_{s_i} \right)
    \end{split}
\end{align}
Here, $\mA_{s_i :} / A_{s_i}$ is the $s_i$'th row of $\mA$ that is normalized to sum up to 1.

The likelihood of a trajectory $(\vt, \vs)$ for an MJP with parameters $(\vpi, \mA)$ can be computed as
\begin{align*}
\begin{split}
    &p(\vt, \vs | \vpi, \mA) = \pi_{s_1}
    \left(\prod_{i=1}^{N} A_{s_{i-1} s_i}\right) 
    \exp\left(-\sum_{i=1}^{N+1} (t_i - t_{i-1}) \sum_{l=1}^{K} A_{s_i l} \right)
    \intertext{We reformulate this expression using indicators $\vone(\cdot)$, which will make the ELBO computation easier}
    &= \left(\prod_{k=1}^{K} \pi_k^{\vone(s_1 = k)}\right)
    \left(\prod_{i=1}^{N} \prod_{k=1}^K \prod_{l=1}^K A_{kl}^{\vone(s_i = k, s_{i+1} = l)}\right)
    \exp\left(-\sum_{i=1}^{N+1} (t_i - t_{i-1}) \sum_{k=1}^{K} \vone(s_i = k) \left(\sum_{l=1}^{K} A_{k l}\right) \right)
\end{split}
\end{align*}
By applying the logarithm to the above equation, we obtain
\begin{align}
\begin{split}
\log p(\vt, \vs | \vpi, \mA) =& 
    \left(\sum_{k=1}^{K} \vone(s_1 = k) \log \pi_k\right)
    + \left(\sum_{i=1}^{N} \sum_{k=1}^K \sum_{l=1}^K \vone(s_i = k, s_{i+1} = l) \log A_{kl}\right)\\
    &- \left(\sum_{i=1}^{N+1} (t_i - t_{i-1}) \sum_{k=1}^{K} \vone(s_i = k) \left(\sum_{l=1}^{K} A_{k l}\right)\right)
\end{split}
    \label{eq:traj-like}
\end{align}

\textbf{Markov modulated Poisson process.}
Distribution of the observations $\vo = (o_1, ..., o_M)$ of an MMPP depends on the latent MJP trajectory $(\vt, \vs)$ and the rates of each state $\vlambda \in \R_+^K$.
The observations $\vo$ are sampled from an inhomogeneous Poisson process 
with piecewise-constant intensity that depends on the current state: $\lambda(t) = \lambda_{s(t)}$.

Likelihood of the observations $\obs$ given $(\vt, \vs)$ and $\vlambda$ can be computed as
\begin{align*}
    p(\obs | \vt, \vs, \vlambda) 
    &= \left(\prod_{i=1}^{N+1} \lambda_{s_i}^{M_{[t_{i-1}, t_i)}}\right)
    \exp\left(-\sum_{i=1}^{N+1} (t_i - t_{i-1})\lambda_{s_i}\right)
    \intertext{where $M_{[t_{i-1}, t_i)}$ is the number of events $o_j$ in the interval $[t_{i-1}, t_{i})$. Again, using indicator functions, we rewrite it as}
    &= \left(\prod_{i=1}^{N+1} \prod_{j=1}^M \left(\prod_{k=1}^{K} \lambda_{k}^{\vone(s_i = k)}\right)^{\vone(o_j \in [t_{i-1}, t_i))}\right)
    \exp\left(-\sum_{i=1}^{N+1} (t_i - t_{i-1})\sum_{k=1}^{K}\vone(s_i = k) \lambda_{k}\right)
\end{align*}
By applying the logarithm, we obtain
\begin{align}
    \begin{split}
    \log p(\obs | \vt, \vs, \vlambda)
    =& \left(\sum_{i=1}^{N+1} \sum_{j=1}^{M} \vone(o_j \in [t_{i-1}, t_i)
    \sum_{k=1}^{K} \vone(s_i = k) \log \lambda_k\right)\\
    &-\left(\sum_{i=1}^{N+1} (t_i - t_{i-1})\sum_{k=1}^{K}\vone(s_i = k) \lambda_{k}\right)
    \end{split}
    \label{eq:obs-like}
\end{align}

\subsection{Derivation of the ELBO}
\label{app:vi-elbo}
The true posterior of an MMPP $p(\vt, \vs | \obs, \vpi, \mA, \vlambda) \propto p(\vt, \vs, \obs | \vpi, \mA, \vlambda) = p(\vt, \vs| \vpi, \mA) p(\obs | \vt, \vs, \vlambda)$ is intractable.
We approximate it with a variational distribution $q(\vt, \vs)$ by maximizing the evidence lower bound (ELBO) \citep{zhang2018advances}
\begin{align*}
    \elbo(q, \vtheta) &= \E_{q(\vt, \vs)} \bigg[\underbrace{\log p(\vt, \vs | \vpi, \mA)}_{\text{trajectory log-likelihood}} + \underbrace{\log p(\obs | \vt, \vs, \vlambda)}_{\text{observations log-likelihood}} \underbrace{- \log q(\vt, \vs)}_{\text{entropy}}\bigg]
\end{align*}
Recall that we model the approximate posterior as $q(\vt, \vs) = q(\vt) q(\vs | \vt)$, where $q(\vt)$ is defined using \name and $q(\vs | \vt)$ is evaluated exactly for each Monte Carlo sample $\vt$.
We rewrite the ELBO as
\begin{align*}
    \elbo(q, \vtheta) &= \E_{q(\vt)} \left[ \E_{q(\vs | \vt)} \bigg[\log p(\vt, \vs | \vpi, \mA) + \log p(\obs | \vt, \vs, \vlambda)- \log q(\vs| \vt)\bigg] - \log q(\vt) \right]
\end{align*}
We already derived the expressions for $\log p(\vt, \vs | \vpi, \mA)$ (\Eqref{eq:traj-like}) and $\log p(\obs | \vt, \vs, \vlambda)$ (\Eqref{eq:obs-like}).
The expression for $\log q(\vs | \vt)$ can be obtained similarly
\begin{align}
    \log q(\vs | \vt) = \sum_{i=1}^{N+1} \sum_{k=1}^{K} \vone(s_i = k) \log q(s_i = k | \vt)
\end{align}
Finally, to compute the log-density $\log q(\vt)$ of a single sample $\vt = (t_1, ..., t_N)$ we use the procedure described in \Appref{app:impl-details}.
We denote $\vz = (z_1, ..., z_N, z_{N+1}) = \mapF(t_1, ..., t_N, T)$, and $z_{-1}$ is the last entry of $\vz$.
\begin{align}
\label{eq:entropy-nondiff}
    \log q(\vt)
    &= \sum_{i=1}^N \log \left|\frac{\partial z_i}{\partial t_i}\right| - z_{-1} 
\end{align}

\textbf{ELBO (non-differentiable version).}
Putting everything together, we get
\begin{align}
    \label{eq:elbo-nondiff}
    \begin{split}
    \operatorname{ELBO}(q, \vtheta) = 
    \E_{q(\vt)} \Bigg[& \E_{q(\vs | \vt)} \bigg[\sum_{k=1}^{K} \vone(s_1 = k) \log \pi_k\\
    &-\sum_{i=1}^{N+1} (t_{i} - t_{i-1}) \sum_{k=1}^{K} \vone(s_i=k) \sum_{l=1}^{K}A_{kl}\\
    &+ \sum_{i=1}^{N} \sum_{k=1}^{K} \sum_{l=1}^{K} \vone(s_i = k, s_{i+1} = l) \log A_{kl}\\
    &+\sum_{i=1}^{N+1} \sum_{j=1}^{M} \vone(o_j \in [t_i, t_{i+1})) \sum_{k=1}^{K} \vone(s_i = k) \log \lambda_k\\
    &-\sum_{i=1}^{N+1} (t_{i} - t_{i-1}) \sum_{k=1}^{K} \vone(s_i = k) \lambda_k\\
    &- \sum_{i=1}^{N+1} \sum_{k=1}^{K} \vone(s_i = k) \log q(s_i = k| \vt)
    \bigg] \\ 
    &-\sum_{i=1}^N \log \left|\frac{\partial z_i}{\partial t_i}\right| + z_{-1}
    \Bigg]
    \end{split}
\end{align}
Note that $N$ is the length of the sample $\vt$ generated from $q(\vt)$, so $N$ will take different values for different samples.
We can evaluate the inner expectation w.r.t.\ $q(\vs | \vt)$ by using the following fact
\begin{align}
    \E_{q(\vs|\vt)} \left[ \vone(s_i = k)\right] = q(s_i = k | \vt) & & 
    \E_{q(\vs|\vt)} \left[ \vone(s_i = k, s_{i+1} = l)\right] = q(s_i = k, s_{i+1} = l | \vt)
\end{align}
Recall that we set $q(\vs | \vt)$ to the true posterior $p(\vs | \vt, \vo, \vtheta)$ over states given the jumps (\Eqref{eq:variational-state}).
This allows us to exactly compute both the posterior marginals $q(s_i = k | \vt)$ and the posterior transition probabilities $q(s_i = k, s_{i+1} = l | \vt)$ using the forward-backward algorithm \citep[Equations 7, 8]{wang2013collapsed}.
Therefore, the inner expectation w.r.t.\ $q(\vs | \vt)$ in \Eqref{eq:elbo-nondiff} can be computed analytically.

\textbf{ELBO (differentiable relaxation).}
The ELBO, as defined above in \Eqref{eq:elbo-nondiff}, is discontinuous 
w.r.t.\ the parameters of the density $q(\vt)$ for the reasons described in \Secref{sec:diff-entropy}.
The expression inside the expectation depends only on the events $t_i$ that happen before $T$.
Infinitesimal change in the parameters of $q(\vt)$ may "push" the point $t_i$ outside $[0, T]$, thus changing the function value by a fixed amount and resulting in a discontinuity.

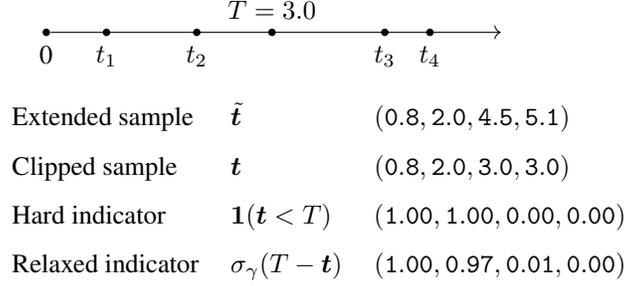
\begin{figure}
    \centering
    \hspace{-15mm}
\begin{tikzpicture}[dot/.style={circle,inner sep=1pt,fill,name=#1},
    extended line/.style={shorten >= -#1,shorten <= 0},
    extended line/.default=1cm,]

    \node [dot=0,label=below:0] at (0, 0) {};
    \node [dot=t1,label=below:$t_1$] at (0.8, 0) {};
    \node [dot=t2,label=below:$t_2$] at (2.0, 0) {};
    \node [dot=t3,label=below:$t_3$] at (4.5, 0) {};
    \node [dot=t4,label=below:$t_4$] at (5.1, 0) {};

    \node [circle,inner sep=1pt,fill,name=T,label={$T=3.0$}] at (3.0, 0) {};

    \draw [extended line=1cm,->] (0) -- (t4);
\end{tikzpicture}

\begin{center}
\begin{tabular}{lll}
    \text{Extended sample} & $\ext{\vt} $ & $\mathtt{(0.8, 2.0, 4.5, 5.1)}$\\
    \text{Clipped sample} & $\clip{\vt} $ & $\mathtt{(0.8, 2.0, 3.0, 3.0)}$\\
    \text{Hard indicator} & $\vone (\vt < T) $ & $\mathtt{(1.00, 1.00, 0.00, 0.00)}$\\
    \text{Relaxed indicator} & $\sigma_\temp(T - \vt)$ & $\mathtt{(1.00, 0.97, 0.01, 0.00)}$\\
\end{tabular}
\end{center}
    \caption{Examples of values involved in the ELBO computation.}
    \label{fig:elbo-terms}
\end{figure}

We fix this problem using the approach described in \Appref{app:impl-details} and \Secref{sec:diff-entropy}.
We obtain an "extended" sample $\ext{\vt}$ by first simulating a sequence $\ext{\vz} = (\ext{z}_1, ..., \ext{z}_{N^\prime})$ from a HPP with unit rate and computing $\ext{\vt} = \mapF^{-1}(\ext{\vz})$ (more on this in \Appref{app:impl-details}).
We get a "clipped" / "padded" sample $\clip{\vt} = (\clip{t}_1, ..., \clip{t}_{N^\prime})$  as $\clip{t}_i = \min\{\ext{t}_i, T\}$ (\Figref{fig:elbo-terms}).
Finally, we compute $\clip{\vz} = (\clip{z}_1, ..., \clip{z}_{N^\prime}) = \mapF(\clip{\vt})$ (this is necessary for computing the correct cumulative intensity $\CumulativeIntensity(T)$ after clipping).
We can now express the ELBO in terms of the "extended" samples $\ext{\vt}$ and "clipped" samples $\vt$:
\begin{align}
    \label{eq:elbo-diff}
    \begin{split}
    \operatorname{ELBO}(q, \vtheta) = 
    \E_{q(\vt)} \Bigg[& \E_{q(\vs | \vt)} \bigg[\sum_{k=1}^{K} \vone(s_1 = k) \log \pi_k\\
    &-\sum_{i=1}^{\hl{N^\prime}} (\clip{t}_{i} - \clip{t}_{i-1}) \sum_{k=1}^{K} \vone(s_i=k) \sum_{l=1}^{K}A_{kl}\\
    &+ \sum_{i=1}^{\hl{N^\prime}} \hl{\vone(\ext{t}_i < T)} \sum_{k=1}^{K} \sum_{l=1}^{K} \vone(s_i = k, s_{i+1} = l) \log A_{kl}\\
    &+\sum_{i=1}^{\hl{N^\prime}} \sum_{j=1}^{M} \vone(o_j \in [t_i, t_{i+1})) \sum_{k=1}^{K} \vone(s_i = k) \log \lambda_k\\
    &-\sum_{i=1}^{\hl{N^\prime}} (\clip{t}_{i} - \clip{t}_{i-1}) \sum_{k=1}^{K} \vone(s_i = k) \lambda_k\\
    &-\sum_{i=1}^{\hl{N^\prime}} \hl{\vone(\ext{t}_{i-1} < T)} \sum_{k=1}^{K} \vone(s_i = k) \log q(s_i = k | \vt)
    \bigg] \\ 
    &-\sum_{i=1}^{\hl{N^\prime}} \hl{\vone(\ext{t}_i < T)} \log \left|\frac{\partial z_i}{\partial t_i}\right| + \clip{z}_{-1}
    \Bigg]
    \end{split}
\end{align}
Changes from \Eqref{eq:elbo-nondiff} are highlighted in \hl{red}.
Even though the formula looks different, the result of evaluating \Eqref{eq:elbo-diff} will be \emph{exactly} the same as for \Eqref{eq:elbo-nondiff}.
By using different notation we only made the process of "discarding" the events $t_i > T$ explicit.
The new formulation allows us to obtain a differentiable relaxation.
For this, we replace the indicator functions $\vone(t_i < T)$ with sigmoids $\sigmoid_\gamma(T - t_i)$.
The indicator function $\vone(o_j \in [t_i, t_{i+1}))$ can also be relaxed as
\begin{align}
\begin{split}
\vone(o_j \in [t_i, t_{i+1})) &= \vone(t_{i+1} > o_j ) - \vone(t_{i} \ge o_j)\\
&\approx \sigma_{\temp}(t_{i+1} - o_j) - \sigma_{\temp}(t_{i} - o_j)
\end{split}
\end{align}
By combining all these facts, we obtain a differentiable relaxation of the ELBO.
Our method leads to an efficient implementation that uses batches of samples.
We sample a batch of jump times $\{\vt^{(1)}, \vt^{(2)}, ...\}$ from $q(\vt)$, evaluate the posterior $q(\vs | \vt)$ using with forward-backward for all of them in parallel, and evaluate the relaxed ELBO (\Eqref{eq:elbo-diff}).

\subsection{Parameter estimation}
\label{app:vi-param-estimation}
In \Secref{sec:diff-mjp}, we perform approximate posterior inference over the trajectories $(\vt, \vs)$ by maximizing the ELBO w.r.t.\ $q(\vt, \vs)$
\begin{align}
    \label{eq:only-posterior}
    \max_{q(\vt, \vs)} \E_q[\log p(\vt, \vs | \vtheta) + \log p(\obs | \vt, \vs, \vtheta) - \log q(\vt, \vs)]
\end{align}
Since $\elbo(q, \vtheta)$ provides a lower bound on the marginal log-likelihood $\log p(\obs | \vtheta)$,
we can also simultaneously learn the model parameters $\vtheta = \{\vpi, \mA, \vlambda \}$ by solving the following optimization problem (subject to appropriate constraints on $\vtheta$)
\begin{align}
    \label{eq:variational-bayes}
    \max_{\vtheta}\max_{q(\vt, \vs)} \E_q[\log p(\vt, \vs | \vtheta) + \log p(\obs | \vt, \vs, \vtheta) - \log q(\vt, \vs)]
\end{align}
Finally, we can perform fully Bayesian treatment and approximate the posterior distribution over the parameters as well as the trajectories.
For this, we can place a prior $p(\vtheta)$ and approximate $p(\vtheta, \vt, \vs | \vx)$ with $q(\vtheta, \vt, \vs) = q(\vtheta) q(\vt) q(\vs | \vt, \vtheta)$.
This corresponds to the following optimization problem
\begin{align}
    \label{eq:fully-bayes}
    \max_{q(\vtheta, \vt, \vs)} \E_q[\log p(\vt, \vs | \vtheta) + \log p(\obs | \vt, \vs, \vtheta) - \log q(\vt, \vs)] - \mathbb{KL}(q(\vtheta) \Vert p(\vtheta))
\end{align}
where $\mathbb{KL}$ denotes KL-divergence.
By applying our relaxation from \Secref{sec:diff}, it's possible to solve all of the above optimization problems (Equations \ref{eq:only-posterior}, \ref{eq:variational-bayes}, \ref{eq:fully-bayes}) using gradient ascent.

\section{Implementation details}
\label{app:impl-details}
\subsection{Batch processing}
By representing TPP densities with transformations, we can implement both (log-)density evaluation and sampling efficiently and in parallel.
Our implementation enables parallelism not only for the events $t_i$ of a single sequence, but also for entire batches consisting of multiple sequences $\vt$ of different length.

First, consider a single sequence $\vt = (1, 2.5, 4)$ with $N = 3$ events, sampled from a TPP on the interval $[0, 5]$.
We pad this sequence with $T = 5$, and additionally introduce a mask $\vm$ that tells us which entries of the padded vector $\vt$ correspond to actual events (i.e., not padding)
\begin{align*}
    \clip{\vt} = \begin{bmatrix}
        1 & 2.5 & 4 & 5
    \end{bmatrix}
    \qquad \qquad
    \vm = \begin{bmatrix}
        1 & 1 & 1 & 0
    \end{bmatrix}
\end{align*}
We implement the transformation $\mapF$ (corresponding to the TPP density $p(\vt)$) similarly to normalizing flow frameworks like \texttt{torch.distributions} \citep{PyTorch}.
We define a method \texttt{forward} that computes $\vz$, the result of the transformation, and $\vj$, logarithm of the diagonal entries of the Jacobian $J_{\mapF}(\vt)$:
\begin{align*}
    \vz = \mapF(\vt) 
    = \begin{bmatrix}
        z_1 & z_2 & z_3 & z_4
    \end{bmatrix}
    \qquad \qquad 
    \vj = \begin{bmatrix}
        \log \left|\frac{\partial z_1}{\partial t_1} \right|
        & \log \left|\frac{\partial z_2}{\partial t_2} \right|
        & \log \left| \frac{\partial z_3}{\partial t_3} \right|
        & \log \left|\frac{\partial z_4}{\partial t_4}\right|
    \end{bmatrix}
\end{align*}
From the definition of $\mapF$ (Table \ref{tab:notation}), we can see that the last entry of $\vz$ (that we denote as $z_{-1}$) corresponds to $\CumulativeIntensity(T)$.
Also, each entry $j_i$ of $\vj$ corresponds to $\log \left|\frac{\partial \CumulativeIntensity(t_i)}{\partial t_i}\right|$.
Therefore, we can compute the log-density $\log p(\vt)$ as 
\begin{align}
    \log p(\vt) = \text{\texttt{sum}}(\vm \odot \vj) - z_{-1} = 
    \sum_{i=1}^{N^\prime} m_i \log \left|\frac{\partial z_i}{\partial t_i}\right| - z_{-1} =
    \sum_{i=1}^{N} \log \left|\frac{\partial \CumulativeIntensity(t_i)}{\partial t_i}\right| - \CumulativeIntensity(T)
    \label{eq:density-implementation}
\end{align}
where $N^\prime$ denotes the length \emph{with} the padding.
We can verify that this is equal to the logarithm of the TPP density in \Eqref{eq:tpp-likelihood}.
Note that if we use a longer padding, such as
\begin{align*}
    \vt = \begin{bmatrix}
        1 & 2.5 & 4 & 5 & 5 & 5 & 5
    \end{bmatrix}
    \qquad \qquad
    \vm = \begin{bmatrix}
        1 & 1 & 1 & 0 & 0 & 0 & 0
    \end{bmatrix}
\end{align*}
then \Eqref{eq:density-implementation} will still correctly compute the log-likelihood for the sequence.
This observation allows to process multiple sequences $\{\vt^{(1)}, \vt^{(2)}, ...\}$ in a single batch.
We simply pad all the sequences with $T$ up to the length of the longest sequence, stack them into a matrix of shape \texttt{[batch\_size, max\_seq\_len]} and process all of them in parallel.

As described in \Secref{sec:tri-tpp-triangular}, we actually define $\mapF$ by stacking multiple transformations.
We sequentially call the \texttt{forward} method for each transformation in the chain to obtain the final $\vz$, and sum up the log-diagonals of the Jacobians $\vj$ along the way.
Each transformation and its Jacobian can be evaluated in parallel in linear time, making the whole operation efficient.

\subsection{Sampling}
Sampling is implemented similarly.
We start by simulating a vector $\ext{\vz}$ from a homogeneous Poisson process with unit rate.
The length of $\ext{\vz}$ must be "long enough" (more on this later).
We define the method \texttt{inverse} that computes $\ext{\vt} = \mapF^{-1}(\ext{\vz})$.
We obtain a final sample $\vt$ by clipping the entries of $\ext{\vt}$ as $t_i = \min\{\ext{t}_i, T\}$.
If we would like to compute the density of the generated sample $\vt$, we will also need the mask $\vm$ that can be obtained as $m_i = \vone(\ext{t}_i < T)$.
In some use cases, such as entropy maximization (\Secref{sec:diff-entropy}) or variational inference (\Appref{app:vi}),
we need to use a differentiable approximation to the mask $m_i = \sigmoid_\gamma(T - \ext{t}_i)$.
This recovers our relaxation from \Secref{sec:diff-entropy}.

By slightly abusing the notation, we use $N^\prime$ to denote the number of events in our initial HPP sample $\ext{\vz} = (\ext{z}_1, ..., \ext{z}_{N^\prime})$.
$N^\prime$ must be large enough, such that the event $\ext{t}_{N^\prime}$ (corresponding to $\ext{z}_N^\prime$) happens after $T$.
We can easily ensure this by setting $N^\prime$ to some large number (e.g., $100$ or $1000$), and increasing it if for some sample $\ext{t}_{N^\prime}$ is less than $T$.
As we saw in \Figref{fig:scalability}, using larger sequence length leads to no noticeable computational overhead when using GPU.

\subsection{Ensuring that the TPP is valid}
\label{app:valid}
We showed in \Secref{sec:tri-tpp} that every TPP density $p(\vt)$ corresponds to a differentiable increasing triangular map $\mapF$ defined by the compensator $\CumulativeIntensity$.
When directly parametrizing $\mapF$, we need to check one of the two equivalent conditions to ensure that our map $\mapF$ defines a valid temporal point process.

\textbf{Condition 1.} 
The compensator $\Lambda^*(t)$ defined by $\mapF$ must be a continuous function of $t$.
(The compensator is already increasing and piecewise-differentiable since $\mapF$ is increasing and differentiable)

\textbf{Condition 2.} The map $\mapF$ is bijective (invertible) on the space of increasing sequences.
In simple words, we need to ensure that for every increasing sequence $\vz = (z_1, ..., z_N)$ of arbitrary length $N$, there exists a unique increasing sequence $\vt = (t_1, ..., t_N)$, such that $\mapF(\vt) = \vz$.

\subsection{Parametrizing transformations using splines}
Rational quadratic splines used by Durkan et al.\ \citep{durkan2019neural} define a flexible nonlinear function $g: (0, 1) \to (0, 1)$.
When defining our TPP models in \Secref{sec:tri-tpp}, we need to parametrize functions $\Lambda: [0, T] \to \R_+$ and $\Phi: \R_+ \to \R_+$ that operate on domains different from $(0, 1)$.
Moreover, we need to ensure domain compatibility when stacking different transformations, such that the overall transformation $\mapF$ is bijective on the space of increasing sequences (\Appref{app:valid}).

We introduce shortcuts for several helper functions that ensure the domain compatibility
\begin{enumerate}
    \item $\vpsi$ applies the function $\psi(x) = 1 - \exp(-x)$ element-wise, where $\psi: \R_+ \to (0, 1)$
    \item $\vpsi^{-1}$ applies the function $\psi^{-1}(y) = -\log (1 - y)$ element-wise, where $\psi^{-1}: (0, 1) \to \R_+$
    \item $\vsigma$ applies the function $\sigma(x) = 1/(1 + \exp(-x))$ element-wise, where $\sigma: \R \to (0, 1)$
    \item $\vsigma^{-1}$ applies the function $\sigma^{-1}(p) = \log p - \log (1 - p)$ element-wise, where $\sigma^{-1} : (0, 1) \to \R$
    \item $\mG$ applies a rational quadratic spline $g: (0, 1) \to (0, 1)$ element-wise.
\end{enumerate}

We implement the transformation for the modulated renewal process (MRP) as
\begin{align*}
    \mapF = \vpsi^{-1} \circ \mG_2 \circ \vpsi \circ \mD \circ \lambda \mI \circ \mG_1 \circ \frac{1}{T} \mI
\end{align*}
where $\mI$ is the identity matrix.

Similarly, we implement the transformation for TriTPP as
\begin{align*}
    \mapF = \vpsi^{-1} \circ \mG_3 \circ \vsigma \circ \mB_L \circ \dots \circ \mB_1 \circ \vsigma^{-1} \circ \mG_2 \circ \vpsi \circ \mD \circ \lambda \mI \circ \mG_1 \circ \frac{1}{T} \mI
\end{align*}
See the code for more details.

\section{Datasets}
\label{app:datasets}

For each synthetic TPP model from Omi et al.\ \citep[Section 4.1]{omi2019fully}, we sampled 1000 sequences on the interval $[0, 100]$.
This includes the \textbf{Hawkes1}, \textbf{Hawkes2}, \textbf{self-correcting (SC)}, \textbf{inhomogeneous Poisson (IPP)}, \textbf{modulated renewal (MRP)} and \textbf{renewal (RP)} processes.

\textbf{PUBG.}\footnote{\url{https://kaggle.com/skihikingkevin/pubg-match-deaths}} 
Each sequence contains timestamps of the death of players in a game of Player Unknown's Battleground (PUBG). 
We use the first 3001 games from the original dataset.

\textbf{Reddit-Comments.} 
Each sequence consists of the timestamps of the comments in a discussion thread posted within 24 hours of the original submission.
We consider the submissions to the \texttt{/r/askscience} subreddit from 01.01.2018 until 31.12.2019.
If several events happen at the \emph{exact} same time, we only keep a single event.
The posts are filtered to have a score of at least 100.
We collected the data ourselves using the \texttt{pushshift} API.\footnote{\url{https://pushshift.io/}}

\textbf{Reddit-Submissions.} 
Each sequence contains the timestamps of submissions to the \texttt{/r/politics} subreddit within a single day (24 hours).
We consider the period from 01.01.2017 until 31.12.2019.
If several events happen at the \emph{exact} same time, we only keep a single event.
The data is again collected using the \texttt{pushshift} API.

\textbf{Taxi\footnote{\url{https://www.kaggle.com/c/nyc-taxi-trip-duration/data}}} contains the records of taxi pick-ups in New York. We restrict our attention to the south of Manhattan, which corresponds to the points with latitude in the interval (40.700084, 40.707697) and longitude in (-74.019871, -73.999443).

\textbf{Twitter\footnote{\url{https://twitter.com}}} contains the timestamps of the tweets by user 25073877, recorded over several years.

\textbf{Yelp 1 and 2\footnote{\url{https://www.yelp.com/dataset/challenge}}} contain the user check-in times for the McCarran International Airport and for all businesses in the city of Mississauga in 2018, respectively.

Table \ref{tab:data-details} shows the number of sequences, average sequence length and the duration of the $[0, T]$ interval for all the datasets. 
\def\arraystretch{1.0}
\begin{table}[h]
    \centering
    \begin{tabular}{l|rrrr}
    Dataset name & Number of sequences &  Average sequence length & Interval duration\\
    \midrule
	Hawkes 1& 1000& 95.4& 100\\
	Hawkes 2& 1000& 97.2& 100\\
	SC& 1000& 100.2& 100\\
	IPP& 1000& 100.3& 100\\
	MRP& 1000& 98.0& 100\\
	RP& 1000& 109.2& 100\\
	\midrule
	PUBG& 3001& 76.5& 40\\
	Reddit-C& 1356& 295.7& 24\\
	Reddit-S& 1094& 1129.0& 24\\
	Taxi& 182& 98.4& 24\\
	Twitter& 2019& 14.9& 24\\
	Yelp 1& 319& 30.5& 24\\
	Yelp 2& 319& 55.2& 24
	\end{tabular}
	\vspace{2mm}
	\caption{Statistics for the synthetic \& real-world datasets}
	\label{tab:data-details}
\end{table}

\section{Experimental setup}
\label{app:setup}
\subsection{Scalability}
\label{app:setup-scalability}
For both the RNN-based model and \name we used 20 spline knots.
We ran \name with blocks of size $H = 16$ and a total of $L = 4$ block-diagonal layers.
This is the configuration of \name with the \emph{largest} number of parameters that we used across our experiments.
For the RNN model, we used the hidden size of 32.
This is the configuration of the RNN model with the \emph{smallest} number of parameters that we used across our experiments.
We did \emph{not} use JIT compilation for either the RNN model or \name, even though enabling JIT would make \name even faster.
When measuring the sampling time, we disabled the gradient computation with \texttt{torch.no\_grad()}.
To remove outliers for the RNN model, we removed 10 longest runtimes for both models.

\subsection{Density estimation}
\label{app:setup-density}
\textbf{NLL.}
In this experiment, we train all models by minimizing the average negative log-likelihood of the training set $\gD_{\text{train}} = \{\vt^{(1)}, \vt^{(2)}, ...\}$
\begin{align*}
    \min_\vtheta - \frac{1}{|\gD_{\text{train}}|}
    \frac{1}{N_{\text{avg}}}
    \sum_{\vt \in \gD_{\text{train}}} \log p_\vtheta(\vt)
\end{align*}
We normalize the loss by $N_{\text{avg}}$, the average number of events in a sequence in the training set,
in order to obtain values that are at least somewhat comparable across the datasets.
We perform full-batch training since the all the considered datasets easily fit into the GPU memory.
For all models, we use learning rate scheduling: if the training loss does not improve for 100 iterations, the learning rate is halved.
The training is stopped after 5000 epochs or if the validation loss stops improving for 300 epochs, whichever happens first.
We train all models using the parameter configurations reported in \Secref{sec:exp-prediction} and pick the configuration with the best validation loss.

\textbf{MMD.}
We train the models \& tune the hyperparameters using the same procedure as in the NLL experiment.
Then, we compare the distribution $p(\vt)$ learned by each model with the empirical distribution $p^\star(\vt)$ on the hold-out test set by estimating the maximum mean discrepancy (MMD) \citep{gretton2012kernel}.
The MMD between distributions $p$ and $p^\star$ is defined as
\newcommand\cdiff[1]{\left\lVert#1\right\rVert_\star}
\begin{align*}
    \text{MMD}(p, p^\star) &= \E_{\vt, \vt^\prime \sim p}[k(\vt, \vt^\prime)] - 2\E_{\vt \sim p, \vu\sim p^\star}[k(\vt, \vu)] + \E_{\vu, \vu^\prime \sim p^\star}[k(\vu, \vu^\prime)]
\end{align*}
Here, $\vt = (t_1, ..., t_N)$ and $\vu = (u_1, ..., u_{M})$ denote variable-length TPP realizations from different distributions, 
and $k(\cdot, \cdot)$ is a positive semi-definite kernel function that quantifies the similarity between two TPP realizations.
We use the Gaussian kernel
\begin{align*}
    k(\vt, \vu) &= \exp\left(-\frac{d(\vt, \vu)}{2\sigma^2}\right)
\end{align*}
\begin{wrapfigure}{r}{0.4\textwidth}
    \includegraphics[width=0.9\linewidth]{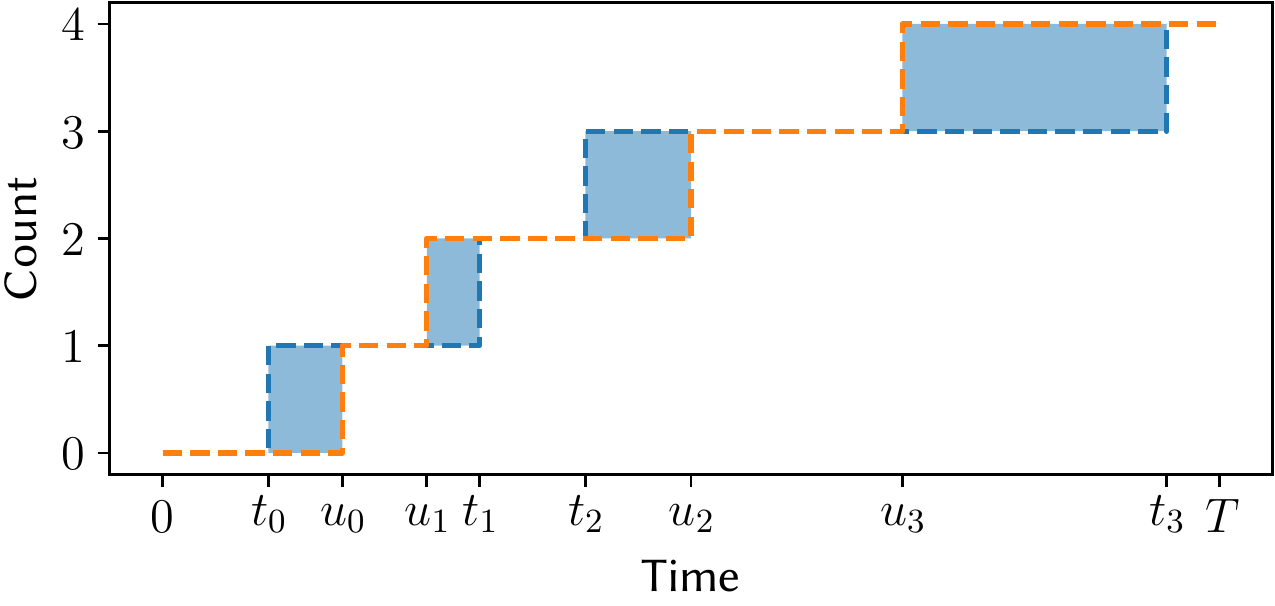}
    \caption{The blue area represents the counting measure distance (figure adapted from \citep{xiao2017wasserstein}).}
    \vspace{-14mm}
\end{wrapfigure}
where $d(\vt, \vu)$ is the counting measure distance between two TPP realizations from \citep[Equation 3]{xiao2017wasserstein}, defined as
\begin{align*}
    d(\vt,\vu) &= \sum_{i=1}^N \left|t_i - u_i\right| + \sum_{i=N+1}^{M}\left(T-u_i\right)
\end{align*}
Here, we assume w.l.o.g.\ that $N \le M$.
Following Section 8 of Gretton et al.\ \cite{gretton2012kernel},
the parameter $\sigma$ is estimated as the median of $d(\vt, \vu)$ with $\vt,\vu\sim p\cup p^\star$.

\subsection{Variational inference}
\label{app:setup-vi}
We simulate an MMPP with $K=3$ states and the following parameters
\begin{align*}
    \mA = \begin{pmatrix}
        0.1 & 0.1 & 0.1\\
        0.1 & 0.1 & 0.1\\
        0.1 & 0.1 & 0.1
    \end{pmatrix}
    &&
    \vpi = \begin{pmatrix}
        0.52\\
        0.22\\
        0.26
    \end{pmatrix}
    &&
    \vlambda = \begin{pmatrix}
        1\\
        5\\
        20
    \end{pmatrix}
\end{align*}
We use the following configuration for \name in this experiment:
$L=2$ blocks of size $H = 4$, learning rate $0.01$, no weight decay.
We estimate the ELBO using 512 Monte Carlo samples from $q(\vt)$ and use the temperature $\gamma = 0.1$ for the relaxation.
We implemented the MCMC sampler by Rao \& Teh \citep{rao2013fast} in Pytorch.
We discard the first 100 samples (burn-in stage), and use 1000 samples to compute the marginal distribution of the posterior.

\section{Additional experiments}
\label{app:additional}
\subsection{Density estimation}
\label{app:additional-density}
\textbf{NLL table with standard deviations.}
For most models \& datasets the results are nearly independent of the random initialization and the standard deviations are very close to zero.
In the following table, we show the standard deviations of the NLL computed over 5 random initializations for all datasets where at least one of the models has the standard deviation above 0.005.
\begin{table}[h]
    \centering
    \begin{tabular}{l|rrrrrrrr}
        {} & PUBG && Reddit-S && Twitter && Yelp2&\\
        \midrule
        {} & mean & std & mean & std & mean & std & mean & std \\ 
        \midrule
        \midrule
        TriTPP& -2.41& 0.34& -4.49& 0.06& 1.06& 0.01& -0.09& 0.01\\
        RNN& -1.97& 0.16& -4.89& 0.29& 1.08& 0.01& -0.07& ---\\
        MRP& -0.83& 0.08& -4.38& 0.05& 1.23& ---& -0.1& ---\\
        RP& 0.12& ---& -4.01& 0.03& 1.2& ---& -0.02& ---\\
        IPP& -0.06& ---& -4.08& ---& 1.61& ---& -0.05& ---\\
    \end{tabular}
\vspace{2mm}
\caption{Average test set NLL with standard deviations. Datasets where all models have a standard deviation below 0.005 are excluded.}
\label{tab:nll-std}
\end{table}

\textbf{Effect of the block size and number on \name performance.}
In this experiment, we show that \name works well with different numbers $L$ and sizes $H$ of block-diagonal layers.
We use the same setup as in the density estimation experiment.
Table \ref{tab:block-size} shows the test set NLL scores for different configurations.
Smaller block are helpful for datasets with a clear global trend (e.g., Reddit-S, Taxi, Yelp), and larger blocks help for datasets with bursty behavior (Reddit-C, Twitter).
In all cases, \name is better than simpler baselines, like MRP, RP and IPP (Table \ref{tab:nll-results}).
\begin{table}[h]
    \resizebox{\columnwidth}{!}{
    \begin{tabular}{l|rrrrrr|rrrrrrr}
        Configuration &  Hawkes1 &  Hawkes2 &    SC &   IPP &   MRP &    RP &  PUBG &  Reddit-C &  Reddit-S &  Taxi &  Twitter &  Yelp1 &  Yelp2 \\
        \midrule
        TriTPP ($L=2,H=4$)& 0.58& 0.01& 0.86& 0.71& 0.35& 0.24& -0.95& -2.26& -4.69& -0.68& 1.11& 0.62& -0.1\\
        TriTPP ($L=4,H=4$)& 0.57& 0.01& 0.85& 0.71& 0.35& 0.24& -2.04& -2.28& -4.57& -0.68& 1.06& 0.63& -0.1\\
        TriTPP ($L=2,H=8$)& 0.56& 0.01& 0.84& 0.71& 0.35& 0.24& -1.93& -2.3& -4.42& -0.66& 1.06& 0.64& -0.09\\
        TriTPP ($L=4,H=8$)& 0.56& 0.0& 0.83& 0.71& 0.35& 0.24& -2.41& -2.33& -4.46& -0.67& 1.06& 0.64& -0.09\\
        TriTPP ($L=2,H=16$)& 0.56& 0.0& 0.84& 0.71& 0.36& 0.25& -1.78& -2.35& -4.45& -0.64& 1.06& 0.67& -0.06\\
        TriTPP ($L=4,H=16$)& 0.56& 0.0& 0.84& 0.72& 0.36& 0.25& -1.83& -2.36& -4.49& -0.64& 1.07& 0.67& -0.06\\
    \end{tabular}
    }
    \vspace{2mm}
    \caption{Test set NLL for different configurations of TriTPP.}
    \label{tab:block-size}
    \vspace{-2mm}
\end{table}

\textbf{Visualizing the effect block-diagonal matrices.}
A completely arbitrary compensator $\CumulativeIntensity$ leads to a completely arbitrary increasing triangular map $\mF$.
However, by picking a parametric class of models, such as MRP or \name, we restrict the set of possible maps $\mapF$ that our model represent.
One way to visualize the dependencies captured by the map $\mapF$ is by looking at its Jacobian $J_\mapF$.

\begin{figure}[h]
    \centering
    \input{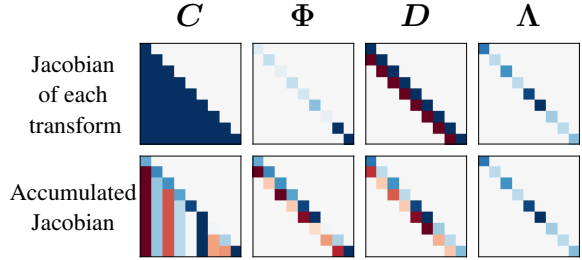}
    \caption{Jacobians of the component transformations of the modulated renewal process.
    We obtain the Jacobian of the combined transformation $\mapF = \mC \circ \mapPh \circ \mD \circ \mapLa$ by multiplying the Jacobians of each transform (right to left).
    }
    \label{fig:jac-mrp}
\end{figure}

\begin{figure}[h]
    \centering
    \resizebox{0.99\textwidth}{!}{\input{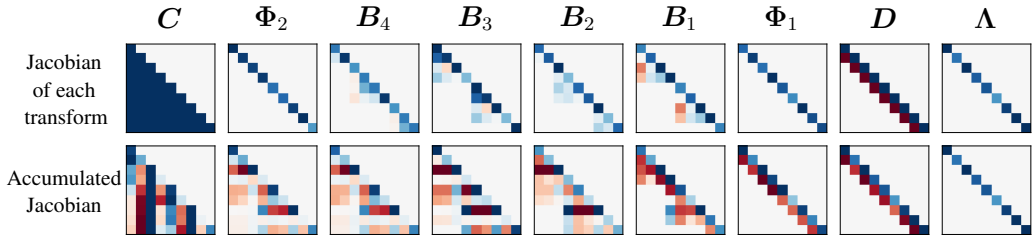}}
    \caption{Jacobians of the component transformations of \name.
    We obtain the Jacobian of the combined transformation $\mapF = \mC \circ \mapPh_2 \circ \mB_4 \circ \mB_3 \circ \mB_2 \circ \mB_1 \circ \mapPh_1 \circ \mD \circ \mapLa$ by multiplying the Jacobians of each transform (right to left). 
    }
    \label{fig:jac-tritpp}
\end{figure}

Figures \ref{fig:jac-mrp} and \ref{fig:jac-tritpp} show the Jacobians of the component transformations for the modulated renewal process and \name.
We can obtain the overall (accumulated) Jacobian of the entire transformation by multiplying the component Jacobians from right to left.
We can see that thanks to the block-diagonal layers \name is able to capture more complex transformations, and thus richer densities, than MRP.
\def\ww{40pt}

\begin{figure}[h]
    \begin{subfigure}{.33\linewidth}
        \includegraphics[width=\linewidth]{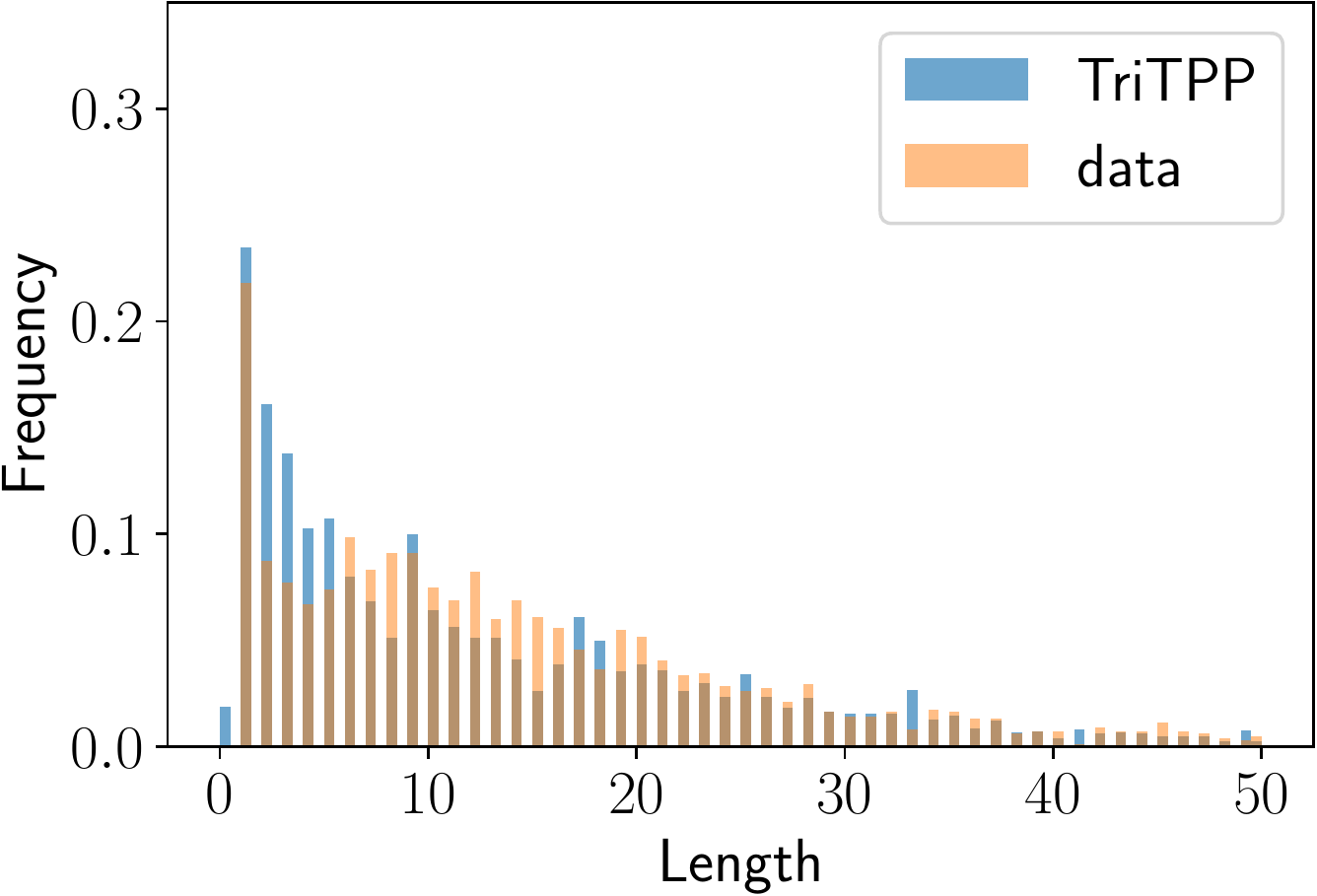}
        \caption{TriTPP (WD $= 0.17$)}
    \end{subfigure}     
    \begin{subfigure}{.33\linewidth}
        \includegraphics[width=\linewidth]{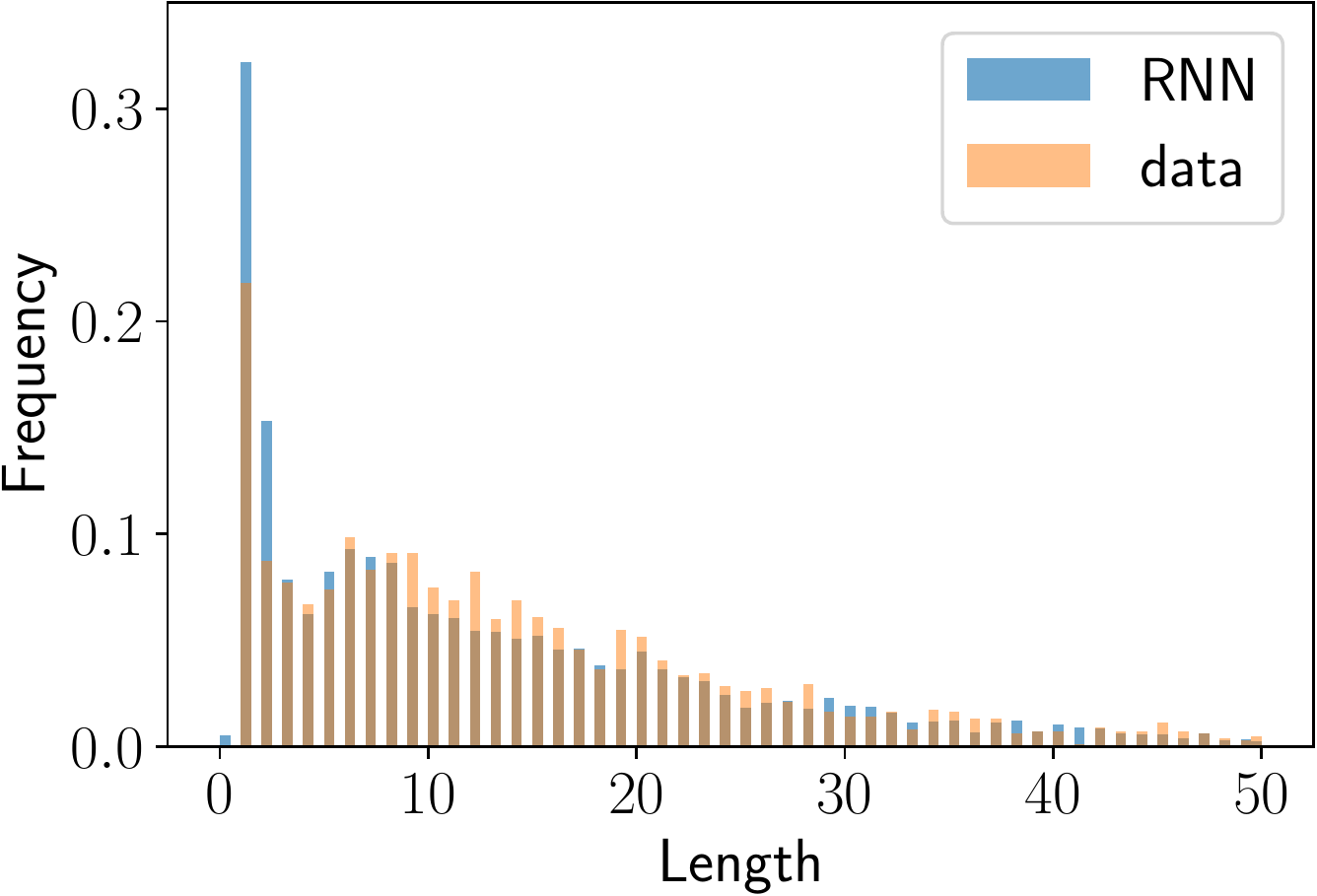}
        \caption{RNN (WD $= 0.10$)}
    \end{subfigure}
    \begin{subfigure}{.33\linewidth}
        \includegraphics[width=\linewidth]{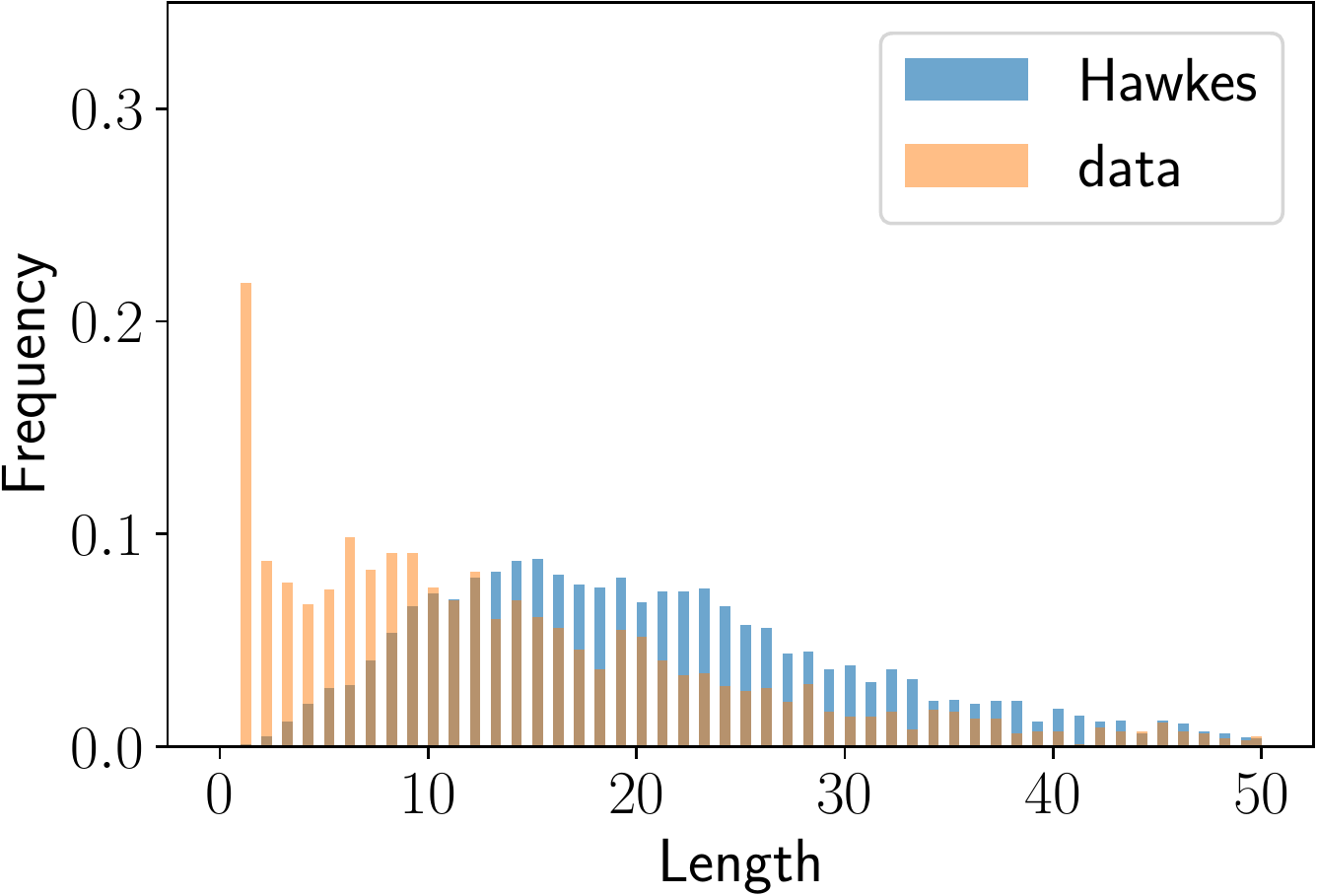}
        \caption{Hawkes (WD $= 0.50$)}
        \label{fig:wasserstein-hawkes}
    \end{subfigure}
    \caption{Histograms of sequence lengths (true and generated) for Twitter. The difference between the two is quantified using Wasserstein distance (WD) --- lower is better.}
    \label{fig:wasserstein}
\end{figure}

\textbf{Distribution of sequence lengths.}
In this experiment, we additionally quantify how well each model captures the true data distribution.
Like before, we train all models on the training set.
We then generate sequences $\vt$ from a trained model and compare the distribution of their lengths to the distribution of the lengths of the true data using Wasserstein distance.
We use the whole dataset since the test sets is too small in some cases.
Using Python pseudocode, this procedure can be expressed as
\begin{align*}
&\text{\texttt{lengths\_sampled = [len(t) for t in model\_samples]}}\\
&\text{\texttt{lengths\_true = [len(t) for t in dataset]}}\\
&\text{\texttt{wd = wasserstein\_distance(lengths\_sampled, lengths\_true)}}
\end{align*}
\Figref{fig:wasserstein} shows the distributions for the Twitter dataset together with the respective Wasserstein distances.
Note that the histograms are used only for visualization purposes, the Wasserstein distance is computed on the raw distributions.
Quantitative results are reported in Table \ref{tab:wasserstein}.
We observe the same trend as before: the RNN-based model and \name consistently outperform the other methods.
Recall that Hawkes process achieves a good NLL on the Twitter data
(Table \ref{tab:nll-results}).
However, when we sample sequences from the trained Hawkes model, the 
distribution of their lengths doesn't actually match the true data, as can be seen in \Figref{fig:wasserstein-hawkes}.

\begin{table}[h]
\resizebox{\textwidth}{!}{
    \begin{tabular}{l|rrrrrr|rrrrrrr}
        {} &  Hawkes1 &  Hawkes2 &    SC &   IPP &   MRP &    RP &  PUBG &  Reddit-C &  Reddit-S &  Taxi &  Twitter &  Yelp1 &  Yelp2 \\
        \midrule
        IPP    &          0.11&         0.11&         0.03&\textbf{0.00}&         0.03&         0.07&\textbf{0.01}&         0.76&         0.27&         0.10&         0.52&         0.07&         0.12 \\
        RP     &          0.07&         0.09&         0.19&         0.14&         0.38&         0.02&         0.67&         0.67&         0.28&         0.85&         0.28&         0.31&         0.20 \\
        MRP    &          0.08&         0.07&         0.02&\textbf{0.00}&\textbf{0.01}&         0.01&\second{0.05}&         0.66&         0.27&         0.09&         0.28&         0.06&\second{0.11} \\
        Hawkes & \second{0.01}&         0.05&         0.03&         0.15&         0.15&         0.03&         0.12&\textbf{0.25}&         0.65&         0.09&         0.50&         0.10&         0.15 \\
        RNN    & \textbf{0.00}&\textbf{0.01}&\textbf{0.00}&         0.04&         0.03&\textbf{0.00}&         0.08&\second{0.40}&\textbf{0.07}&\textbf{0.08}&\textbf{0.10}&\textbf{0.05}&         0.12 \\
        TriTPP &          0.05&\second{0.03}&\textbf{0.00}&         0.01&\textbf{0.01}&\textbf{0.00}&\second{0.05}&         0.53&\second{0.24}&\textbf{0.08}&\second{0.17}&\textbf{0.05}&\textbf{0.09} \\
    \end{tabular}
}
\vspace{2mm}
\caption{Wasserstein distance between the distributions of lengths of true and sampled sequences.}
\label{tab:wasserstein}
\end{table}

\subsection{Variational inference}
\label{app:additional-vi}
\begin{figure}[h]
    \centering
    \includegraphics[width=0.32\textwidth]{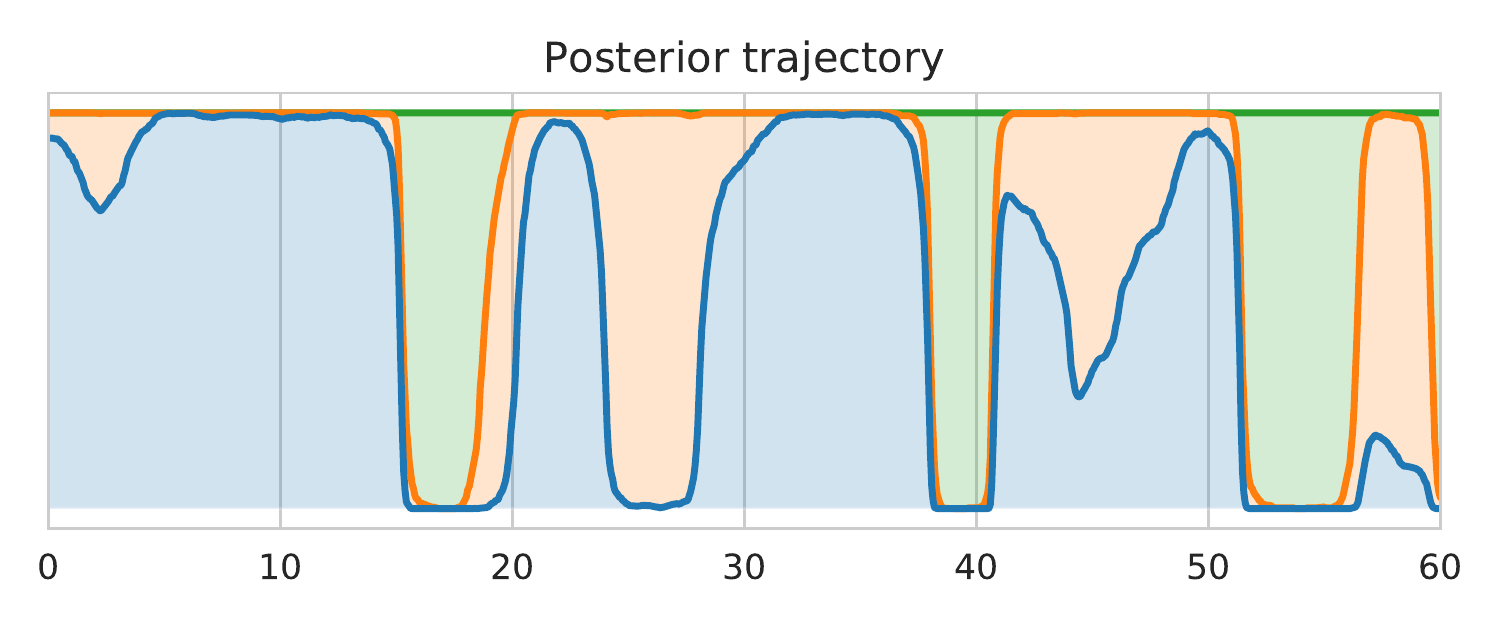}
    \includegraphics[width=0.32\textwidth]{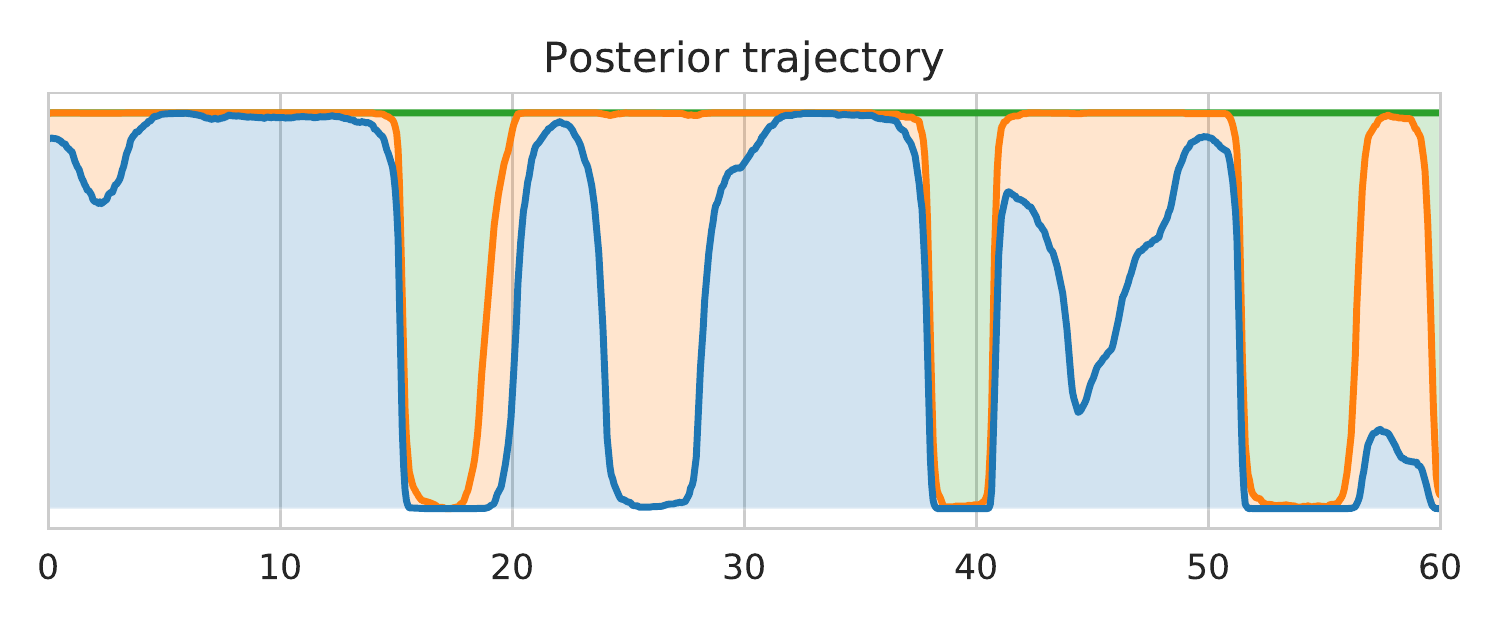}
    \includegraphics[width=0.32\textwidth]{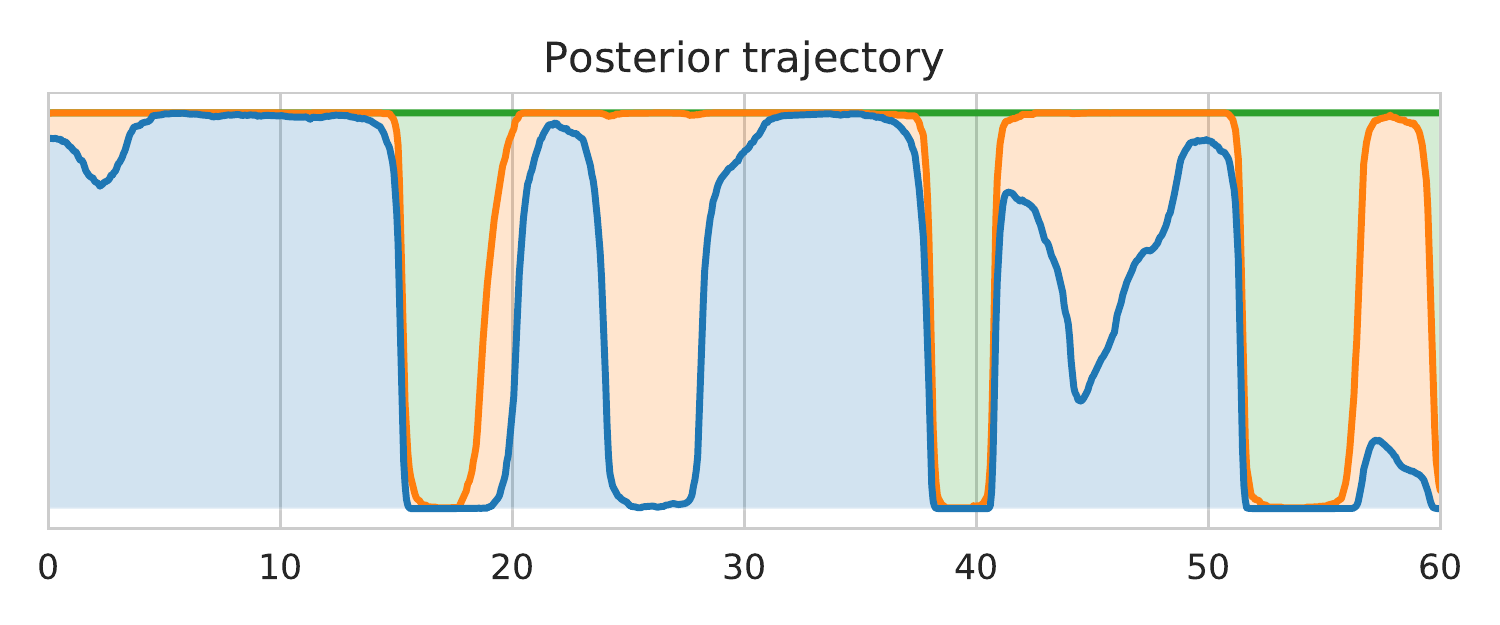}
    \caption{Marginal posterior trajectories obtained when using different random seeds.}
    \label{fig:vi-initializations}
\end{figure}

\textbf{Random initliazations.}
In order to show that ours results are not cherry-picked, we provide the plots of marginal posterior trajectories (similar to \Figref{fig:vi-posterior}) obtained with 3 different random seeds.
\Figref{fig:vi-initializations} shows that our results are consistent across the random seeds.

\begin{figure}[h]
    \centering
    \includegraphics[width=0.5\textwidth]{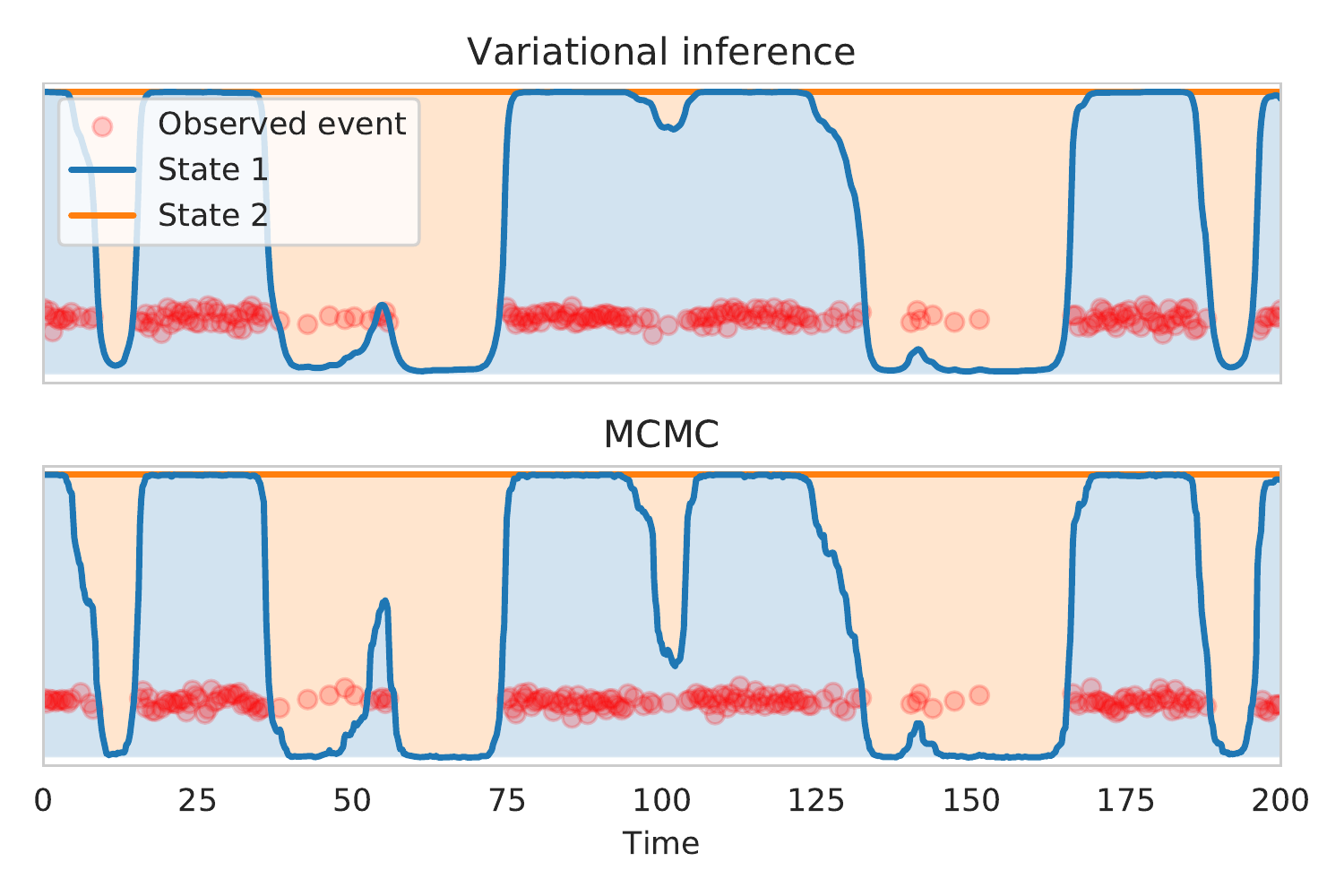}
    \caption{Segmentation of server data obtained using our VI approach and MCMC. In both cases, we estimate the posterior $p(\vt, \vs | \vo, \vtheta)$ as well as the MMPP parameters $\vtheta$.}
    \label{fig:vi-real-world}
\end{figure}

\textbf{Variational inference on real-world data.}
We apply our model to the server log data \footnote{\url{https://www.kaggle.com/shawon10/web-log-dataset}}.
More specifically, we perform segmentation on the interval that contains the first 200 events.
We estimate the posterior over the trajectories $(\vt, \vs)$ and learn the model parameters $\vtheta = \{\vpi, \mA, \vlambda\}$ using the procedure described in \Eqref{eq:variational-bayes}.
Like before, we compare our approach to the MCMC sampler of Rao \& Teh.
For the MCMC sampler, we adopt an EM-like approach, where we alternate between closed-form parameter updates for $\vtheta$ and simulating the posterior trajectories.
\Figref{fig:vi-real-world} shows the obtained posterior trajectories for the two approaches.
Both models learn to segment the sequence into a high-event-rate and a low-event-rate states.

\newpage
\subsection{Miscellaneous}

\textbf{Convergence plots for density estimation.}
\begin{figure}[h]
    \begin{subfigure}{.31\linewidth}
        \includegraphics[width=\linewidth]{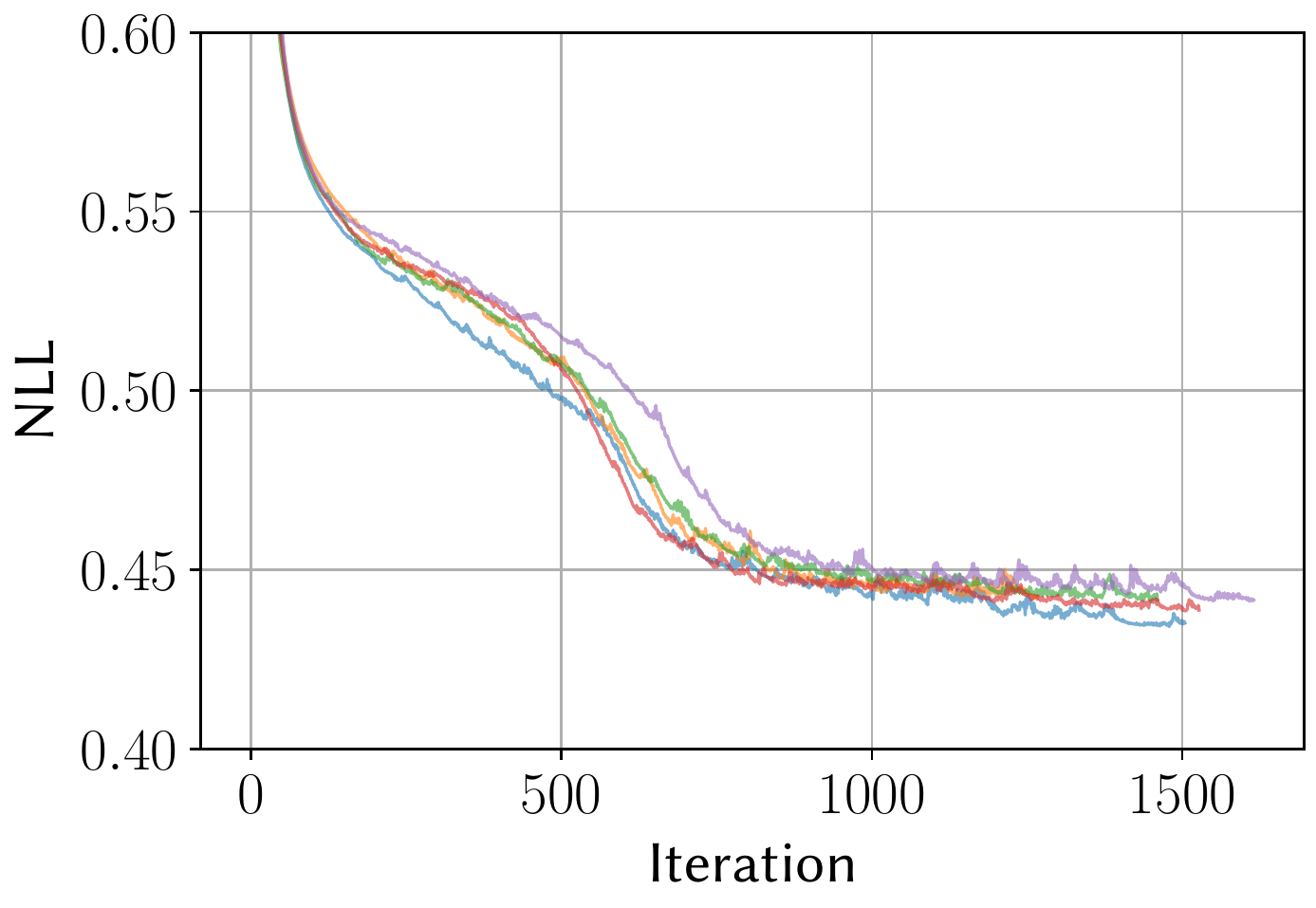}
        \caption{Twitter --- TriTPP}
    \end{subfigure}
    \begin{subfigure}{.31\linewidth}
        \includegraphics[width=\linewidth]{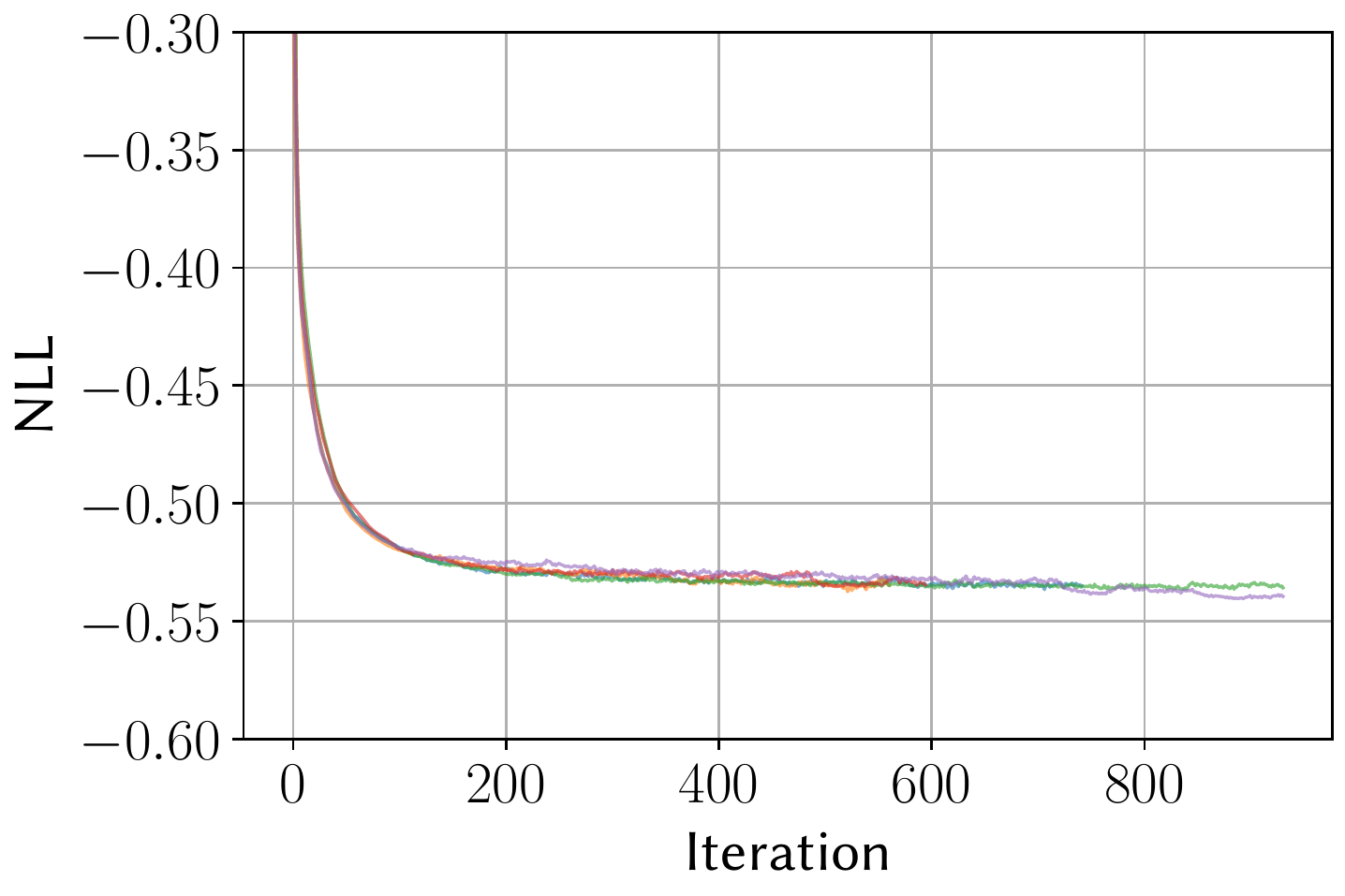}
        \caption{Taxi --- TriTPP}
    \end{subfigure}
    \begin{subfigure}{.31\linewidth}
        \includegraphics[width=\linewidth]{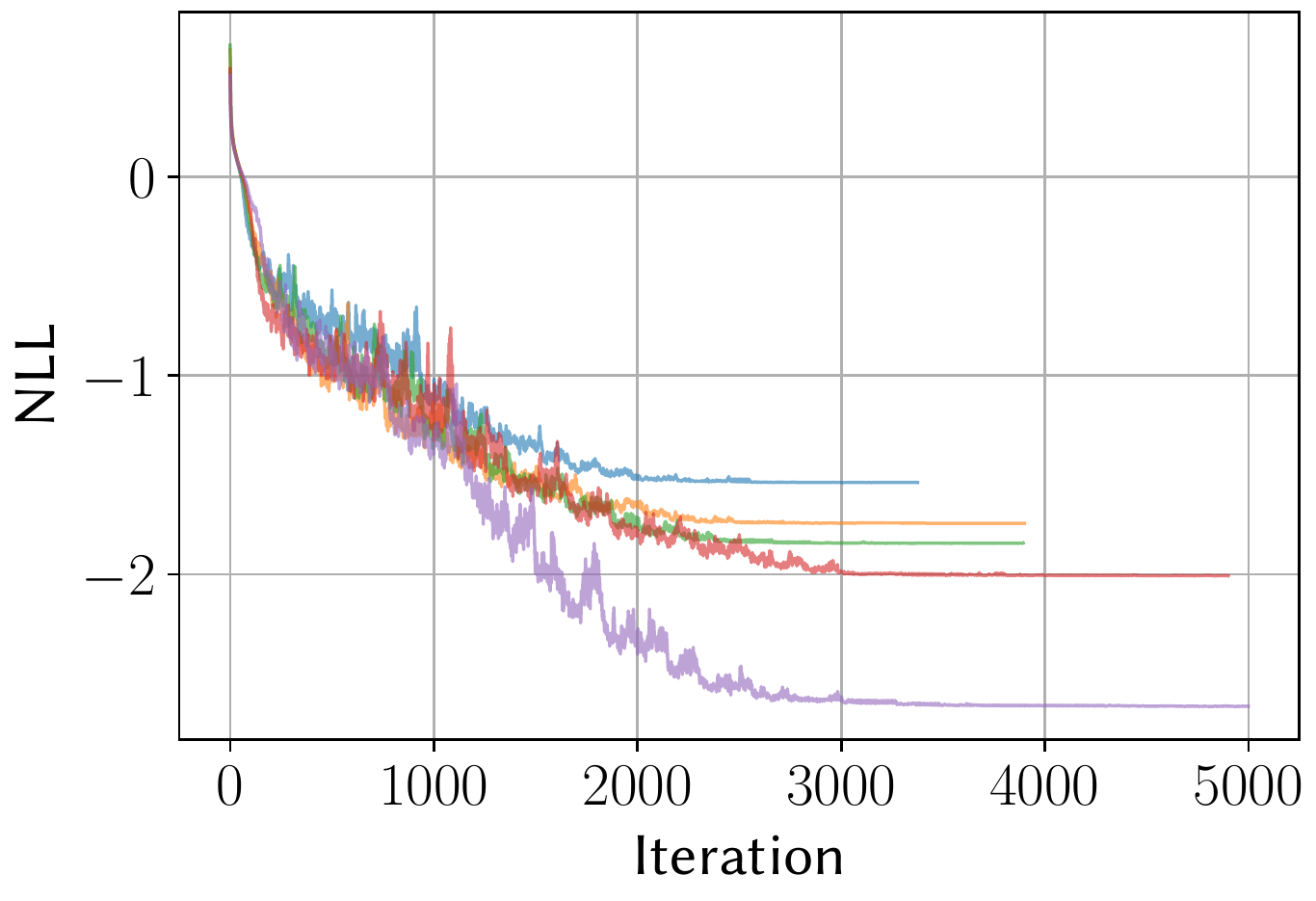}
        \caption{PUBG --- TriTPP}
    \end{subfigure} 

    \begin{subfigure}{.31\linewidth}
        \includegraphics[width=\linewidth]{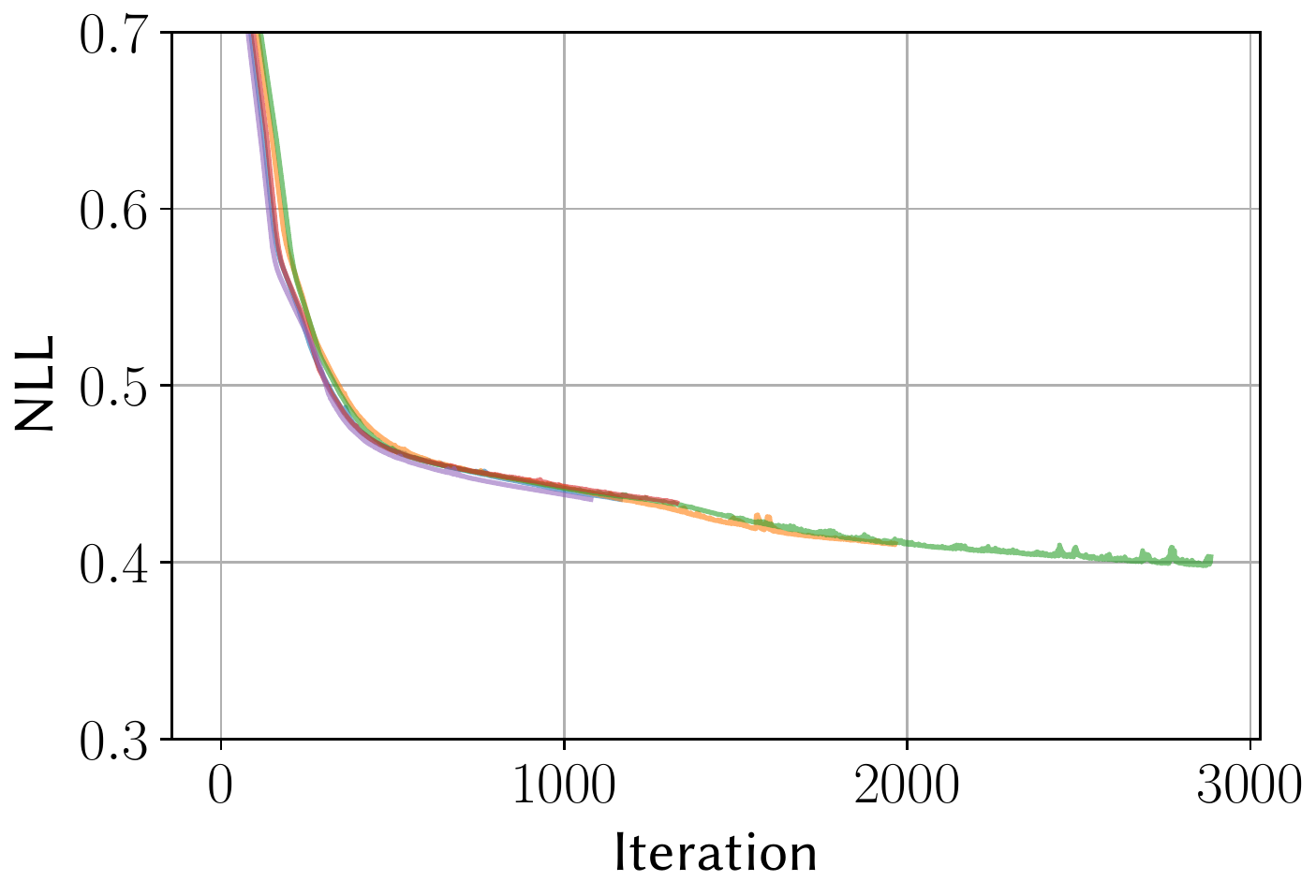}
        \caption{Twitter --- RNN}
    \end{subfigure}
    \begin{subfigure}{.31\linewidth}
        \includegraphics[width=\linewidth]{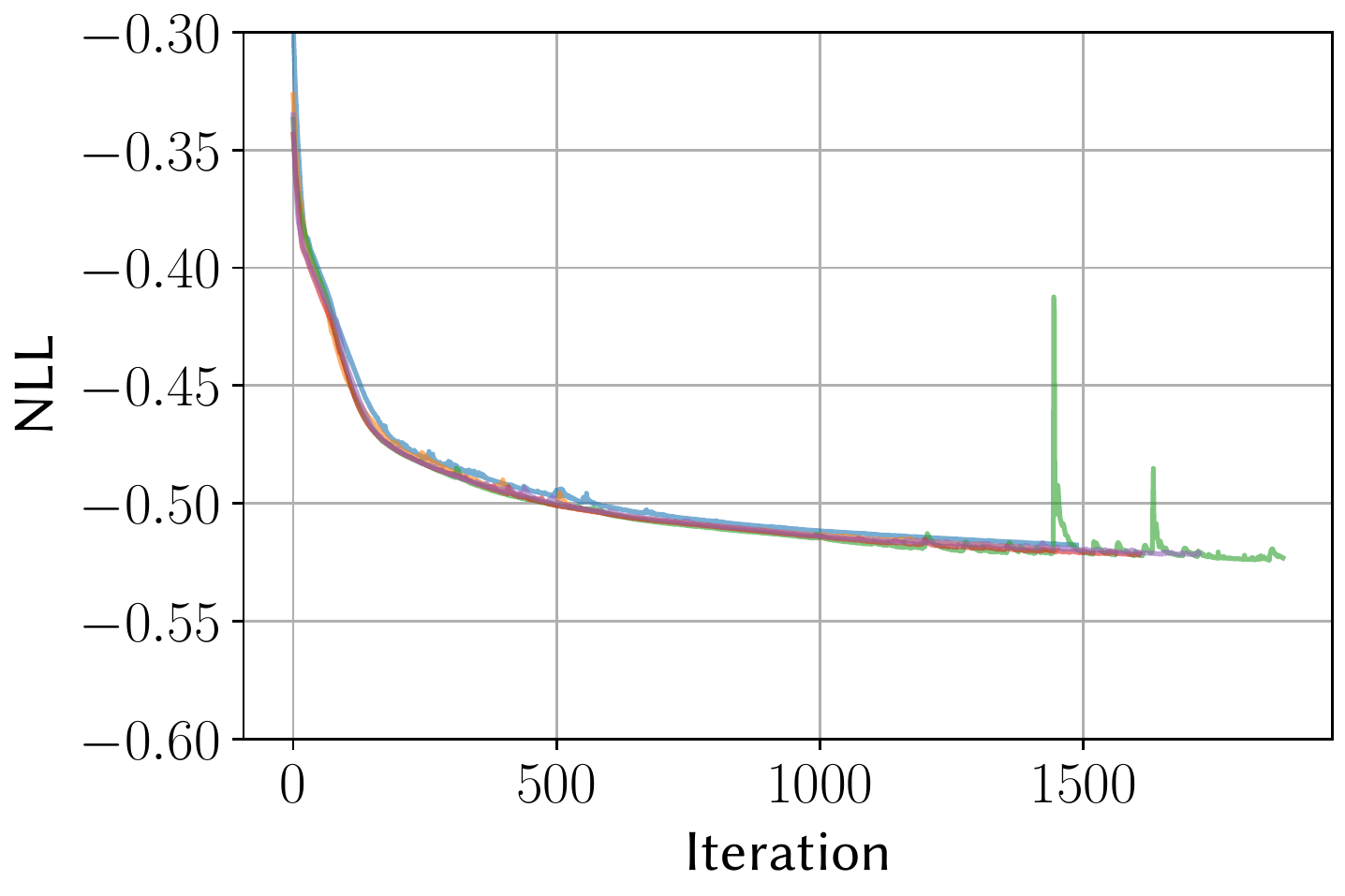}
        \caption{Taxi --- RNN}
    \end{subfigure}
    \begin{subfigure}{.31\linewidth}
        \includegraphics[width=\linewidth]{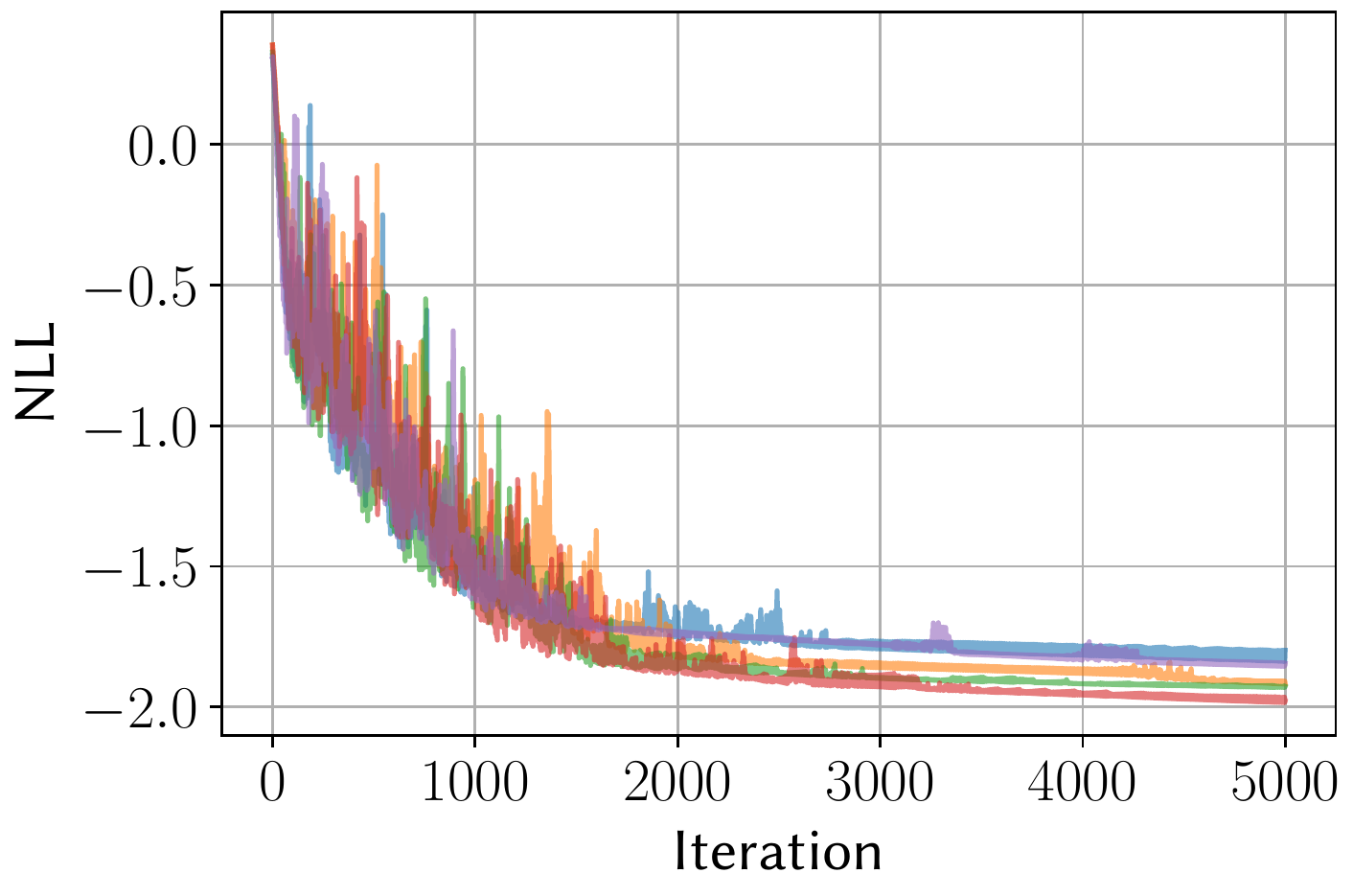}
        \caption{PUBG --- RNN}
    \end{subfigure}
    \caption{Training loss convergence for \name and RNN model with different random seeds.}
\end{figure}

\textbf{Convergence plots for variational inference.}

\begin{figure}[h]
    \centering
    \includegraphics[width=0.45\textwidth]{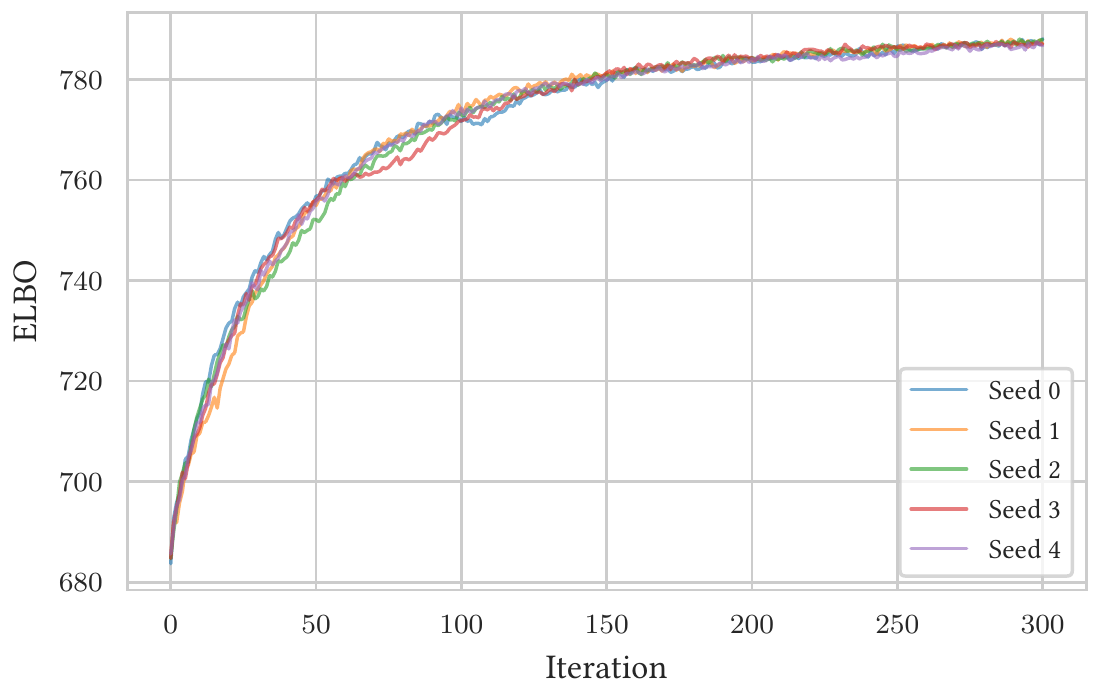}
    \caption{Convergence of our variational inference procedure when using 5 different random seeds.}
\end{figure}

\end{document}